\documentclass[11pt]{article}

\usepackage[preprint]{acl}
\usepackage{wrapfig}
\usepackage{microtype}
\usepackage{hyperref}
\usepackage{url}
\usepackage{booktabs}
\usepackage{adjustbox}
\usepackage{graphicx}
\usepackage{xcolor}
\usepackage[most]{tcolorbox}
\usepackage{enumitem}
\usepackage{amssymb}
\usepackage[most]{tcolorbox}
\usepackage{times}
\usepackage{latexsym}
\usepackage{placeins}

\usepackage{multirow}
\usepackage[T1]{fontenc}
\usepackage[utf8]{inputenc}
\usepackage{latexml}

\usepackage{microtype}

\usepackage{inconsolata}

\usepackage{graphicx}

\title{Spatial Reasoning via Modality Switching Between Language and Symbolic Representations}

\iflatexml
\author{
\mbox{Shreya Rajpal \enspace Tanawan Premsri \enspace Parisa Kordjamshidi} \\
Department of Computer Science and Engineering \\
Michigan State University \\
Michigan, USA \\
\texttt{\{rajpalsh, premsrit, kordjams\}@msu.edu}
}
\else
\author{
Shreya Rajpal \quad Tanawan Premsri \quad Parisa Kordjamshidi \\
Department of Computer Science and Engineering \\
Michigan State University \\
Michigan, USA \\
\texttt{\{rajpalsh, premsrit, kordjams\}@msu.edu}
}
\fi

\begin{document}
\maketitle
\begin{abstract}
Human reasoning is inherently multimodal: when problems become difficult, we rarely think in words alone.
We often externalize our reasoning by sketching diagrams or drawing grids to understand the underlying conceptual structure and avoid mistakes. Building on this premise, our research investigates: (a) whether grounding multi-hop textual-spatial stories into geometry-aware modalities, such as layouts or grids, improves reasoning compared to natural language-based inference; and (b) whether a model can decide when to rely on natural language reasoning and when to switch to a structured modality.
We address these questions by introducing a switching metric based on trustworthiness and complexity signals, which estimates when grounding a spatial story into structure is likely to improve performance. This takes a first step toward principled modality selection in Large Language Model (LLM) reasoning. Across our settings, switching from natural language-based reasoning to a grid-based representation improves LLM performance by up to 42\%, highlighting the importance of modality choice in shaping reasoning outcomes.
\end{abstract}

\section{Introduction}
 Spatial reasoning is becoming increasingly important across many research domains,
 including autonomous driving, navigation, and robotics~\citep{vlmautonomous, tagmap, robospatial,ma2026clamp, ma26breaking}. However, LLMs still exhibit significant limitations in spatial reasoning, especially as spatial context becomes more complex~\citep{survey}. Such tasks require models to compose multiple relations while maintaining a consistent understanding of where entities are located relative to one another. In multi-hop settings, these relations become harder to track, leading to accumulated errors and inconsistent spatial interpretations~\citep{stepgameorig, nesytan}. Even strong LLMs struggle on benchmarks like
 StepGame~\citep{stepgameorig}, SpartQA~\citep{spartqa}, and ReSQ~\citep{spartun} without specialized prompting or step-by-step reasoning traces~\citep{resq, sparc, pathofthought}. However, these methods still rely largely on natural language reasoning, forcing models to infer geometry and layout implicitly from text.

\begin{figure*}[t]
    \vspace{-4pt}
    \centering
    \includegraphics[width=.82\linewidth,height=.25\textheight,keepaspectratio]{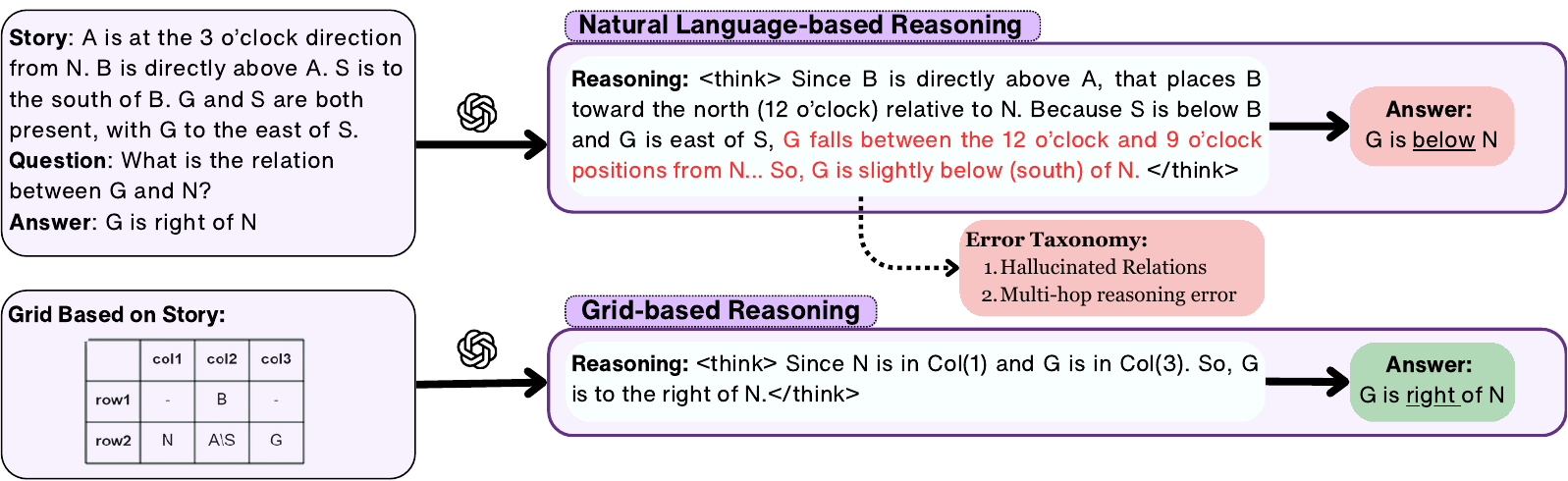}
    \caption{Natural language vs. grid-based reasoning in a multi-hop spatial setting.}
    \label{fig:intro} 
    \vspace{-8pt}
\end{figure*}

This contrasts with human reasoning. For complex spatial descriptions, people often use sketches, diagrams, or grids to externalize the scene into a simplified structure rather than reasoning in words alone~\citep{uno,rexigel2024multiple_external_representations_meta}. This reflects the idea of schematization, which makes spatial structure easier to perceive and reason over~\citep{talmy2003toward}. Similarly, Figure~\ref{fig:intro} shows that natural language-based LLM reasoning must track relations implicitly across multiple steps, which can cause hallucinated relations or multi-hop errors~\citep{sparc}, while a grid makes the structure explicit. This motivates a central question, that is, can spatial representations that explicitly encode geometric structure serve as a more effective reasoning medium for language models than natural language alone?

To investigate this, we compare reasoning directly from natural language stories against different reasoning modalities. Here, \emph{modality} refers to the text-based representation of a spatial story used for inference. These alternatives explicitly encode the scene’s spatial structure, including extracted relational triples and layout-based forms. \footnote{Code is available at:
\url{https://github.com/HLR/Spatial-Modality-Switching}}
As illustrated in the example in Figure~\ref{fig:intro}, explicit structure (e.g., grid-based layouts) can offer a more effective medium for multi-hop spatial reasoning than text alone. Based on this intuition, we propose a grid-grounding framework that converts spatial stories into 2D grids while preserving 3D spatial relations between entities through symbolic encoding. The resulting grids are then fed back to the language model for multi-hop spatial question answering.

However, reasoning with structured representations is not always necessary; in some cases, natural language-based reasoning can be sufficient depending on model ability and problem complexity. Therefore, we introduce a switching metric that estimates when a model should rely on natural language and when it should use structured spatial representations. The metric combines trustworthiness and complexity signals from each instance to enable principled modality selection rather than reasoning with a fixed modality.

Across multiple spatial reasoning benchmarks, our framework improves accuracy by large margins: up to 42\% on StepGame~\citep{stepgameimproved}, 8\% on SpaRTUN~\citep{spartun}, and 5\% on ReSQ~\citep{spartun}. These results highlight the importance of treating visualization and structured symbolic representations not just as outputs but as a reasoning medium. 

In summary, our contributions are:
(1) We investigate whether changing the reasoning modality from natural language to explicit spatial structure (e.g., grid-based representation) improves multi-hop spatial reasoning. 
(2) We propose an adaptive switching framework that uses trustworthiness and complexity signals to decide when a model should reason with natural language and when it should switch to a structured representation, taking a step toward principled modality selection in LLM’s reasoning.

\section{Related Work}
Recent studies show that language models struggle with multi-hop spatial reasoning across benchmarks such as StepGame~\citep{stepgameorig, stepgameimproved}, SpartQA~\citep{spartqa}, SpaRTUN~\citep{spartun}, ReSQ~\citep{spartun}, and SpaRP~\citep{sparc}. Even strong models remain brittle in spatial inference, often making
elementary reasoning errors, failing to compose spatial relations reliably across various spatial
settings~\citep{aggpt4,agdiale,sparc,forest,saturn}.

Prior work has addressed these failures through two broad directions. 
One line of work incorporates symbolic or structured reasoning components to
support spatial reasoning, including neuro-symbolic
frameworks~\citep{nesycoco,nesyframeworks},
methods that combine LLM-based relation extraction with Answer Set Programming~\citep{llmasp, empasp, dspy} or Prolog~\citep{resq}, enforcing logical constraints during fine-tuning~\citep{nesytan}, or using graph neural networks to model multi-step spatial dependencies~\citep{depwiggn, graph}. 
These methods often require additional training or external solvers to improve spatial reasoning. 
In contrast, we study how the representation provided to the language model itself affects spatial reasoning at inference time.

A closer line of work studies whether spatial reasoning improves when intermediate reasoning is represented differently. Tree-of-Thought prompting explores multiple reasoning paths for multi-hop spatial inference~\citep{stepgameimproved}, while Chain-of-Symbol~\citep{chainofsymbol} and coordinate-based formulations~\citep{coordinate} show that symbolic or quantitative representations can improve spatial reasoning. Our work aligns with this direction but compares multiple representations, including natural language, relational triples and coordinate layouts, and introduces grid grounding as a geometry-preserving representation for LLM reasoning.

\section{Methodology} 
\begin{figure*}[t]
    \vspace{-4pt}
    \centering
    \includegraphics[width=.58\textwidth,height=.18\textheight,keepaspectratio]{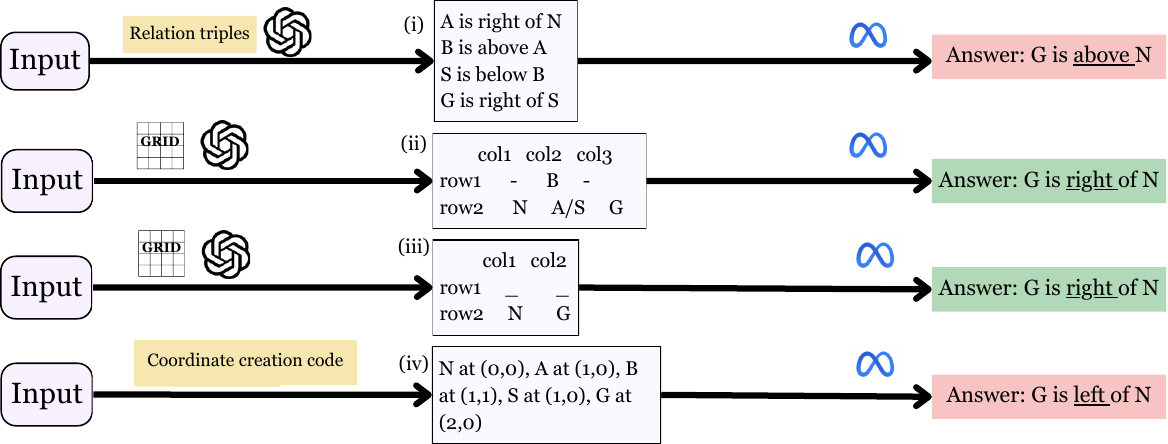}
    \caption{Mapping the story and question from Figure~\ref{fig:intro} into four reasoning representations: (i) relational triples, (ii) full grid, (iii) pruned grid, and (iv) coordinate-based representation. Each modality is used as a reasoning medium for the same question.}
    \label{fig:reasoning-pipelines}
    \vspace{-8pt}
\end{figure*}
\begin{figure*}[t]
    \vspace{-3pt}
    \centering
    \includegraphics[width=.7\linewidth,height=.30\textheight,keepaspectratio]{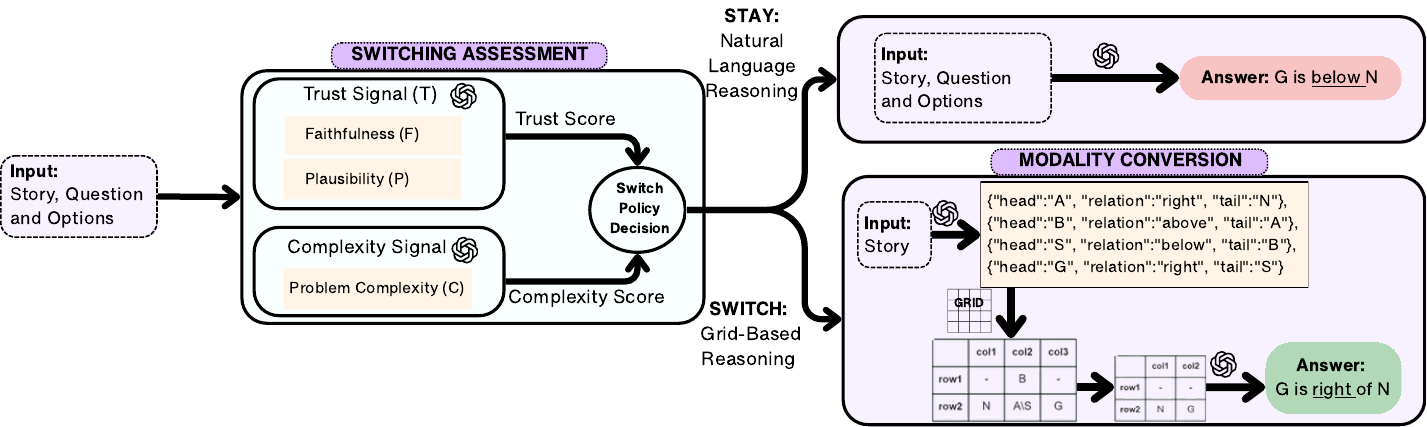}
    \caption{Overview of the proposed pipeline and switching mechanism, using the same story and question input of Figure~\ref{fig:intro}.}
    \label{fig:pipeline}
    \vspace{-5pt}
\end{figure*}

\paragraph{Problem Definition:} We consider a spatial question answering task where, given a spatial narrative ($S$) and a question ($Q$), the model answers the question by inferring the spatial relations between two or more entities. The answer can be either in Yes/No form or as a multi-class label from a fixed set of spatial relations, such as \textit{left} and \textit{above}.
This is a challenging problem because relations between entities can be implicit, and the model often needs to reason across multiple relations to infer the correct answer.

To address this problem, we hypothesize that, depending on the context, some modalities can simplify reasoning more effectively than natural language.
Figure~\ref{fig:intro} illustrates this with a multi-hop story where inferring the relation between $G$ and $N$ requires tracking several intermediate entities in text and can be complex, while the grid makes the relevant relation more clear. 
Therefore, we design a framework to analyze the input and model output and switch between text and structured modalities when needed. 
The overall pipeline is shown in Figure~\ref{fig:pipeline}. It has two main components: (i) assessing problem complexity and model trustworthiness, and (ii) converting the narrative into alternative reasoning modalities.
The assessment module identifies cases where natural language-based reasoning is unreliable or the narrative is too complex and triggers a switch to a structured modality.

Building upon this pipeline, we examine relational triples, coordinate-based layouts, and grids as alternative representations for reasoning and study how switching from natural language-based reasoning to a suitable structured modality affects model performance. We next describe the reasoning modalities and switching policy.

\subsection{Mapping to Multiple Reasoning Modalities}

We transform each spatial narrative and question into multiple qualitative and quantitative representations. Qualitative representations capture symbolic relations, such as relative direction or containment, while quantitative representations make spatial arrangements explicit through coordinate-based layouts. Grid-based representations lie between these two by organizing entities in a discrete row-column layout. Each modality is then used independently for reasoning. Figure~\ref{fig:reasoning-pipelines} illustrates the modalities considered in our framework, discussed as follows.

\noindent\textbf{1. Original Text:}  This is the baseline where the model reasons directly over the original story, question, and answer options in natural language form.

\noindent\textbf{2. Relational Triples:} 
For this modality, we use few-shot in-context learning to extract pairwise spatial relations from the story as triples of the form \texttt{\{head, relation, tail\}}. This process converts the narrative into a qualitative spatial representation, especially when a sentence contains multiple relations or requires coreference resolution. For example, from "a medium purple object is inside and touching block H," we extract \texttt{\{head: medium purple object, relation: tangential-proper-part, tail: block H\}}. These triples include directional relations, such as \textit{left}, \textit{right}, \textit{above} and \textit{below}, as well as topological relations such as \textit{disconnected}, \textit{externally connected}, \textit{partially overlapping}, \textit{equal}, \textit{tangential proper part}, \textit{non-tangential proper part}, \textit{tangential proper part inverse}, \textit{non-tangential proper part inverse}, when needed~\citep{kordjamshidi-etal-2010-spatial}. As shown in Figure~\ref{fig:reasoning-pipelines} (i), the model receives the relational triples with the question and infers the final answer.

\noindent\textbf{3. Grids:} For this modality, we map extracted spatial relations into a two-dimensional grid that assigns relative positions to entities, while preserving three-dimensional spatial information and topological constraints. Topological and three-dimensional relations are encoded with symbolic tags, such as \texttt{\#dc} for disconnected and \texttt{\#behind} for behind. We generate the grid using a Python program with predefined relation templates that translate extracted relation triples into grid placements. The resulting \emph{full grid} serves as a symbolic textual representation of the complete scene. We also construct the \emph{pruned grid}, which retains only question-relevant entities identified by the model based on the story and question, reducing noisy context and isolating the relations needed for the answer. Figure~\ref{fig:reasoning-pipelines} (ii-iii) shows both grid settings. We provide the model with either the full grid or the pruned grid, containing only question-relevant entities such as $N$ and $G$, together with the question for reasoning.

\noindent\textbf{4. Coordinate-based Layout:}
This is a quantitative representation in which the extracted spatial
relations are translated into constraints over the spatial configuration of
objects. Each extracted relation triple is converted into its corresponding
Prolog predicate, and all predicates for a story are solved jointly as a
constraint-satisfaction problem using Prolog's Constraint Logic Programming
over Finite Domains (CLP(FD)) to find a feasible set of
coordinates~\citep{prolog,clpfd}. For example, \texttt{left(A,B)} requires
the two entities to share the same vertical coordinate and places $A$
one unit to the left of $B$, such that
$x_A+1=x_B,\;y_A=y_B$. Inverse relations are defined by reversing the
arguments, such that
\texttt{right(A,B)} $\Leftrightarrow$ \texttt{left(B,A)}. Accordingly,
jointly solving \texttt{left(X,W)} and \texttt{right(O,W)} can yield the
coordinate assignment
\texttt{X=(0,0), W=(1,0), O=(2,0)}. As shown in
Figure~\ref{fig:reasoning-pipelines} (iv), the resulting relative coordinates,
together with the question, are provided to the model for reasoning. Further implementation details are provided in
Appendix~\ref{app:prolog-coordinate}.

\subsection{Switching Between Modalities}
\label{sec: decision}
Natural language-based reasoning can be effective for some spatial narratives, but its reliability depends on the model’s spatial reasoning ability and the instance complexity~\citep{sparc, aggpt4}. As shown in Figure~\ref{fig:pipeline}, complex stories may require composing several relations across intermediate entities to infer the relation between $G$ and $N$. Natural language-based reasoning must track these relations implicitly, while an effective structured representation, such as a grid, makes them explicit and supports more direct and potentially less error-prone inference.

Based on this, we guide modality choice with 2 factors: \emph{trustworthiness}, which reflects the reliability of the model’s natural language-based spatial reasoning, and \emph{complexity}, which captures the difficulty of the problem. Following~\citet{lext,agarwalfaithfulness}, we decompose trustworthiness into \emph{faithfulness} and \emph{plausibility}, which we adapt to spatial reasoning as described next.

\subsubsection{Faithfulness}

Faithfulness, denoted as $F$, measures whether the model’s answer is based on the supporting sentences it identifies from the spatial narrative and whether its reasoning is consistent with those sentences. For example, in the narrative \textit{"Y is positioned below L and to the right; D is upper right of L"}, a question about the relation of $Y$ to $L$ should rely only on the first sentence. If the model uses irrelevant evidence or derives an unsupported relation, its reasoning is not faithful.
To compute $F$, we ask the model to provide supporting sentences along with its answer and evaluate them using two criteria:

(i) \emph{Sufficiency} tests whether the supporting sentences alone recover the original answer. We provide only the supporting sentences to the model and score 1 if the answer matches and 0 otherwise. 

(ii) \emph{Necessity} tests whether the stated supporting sentences are required to derive the answer. We remove them from the story and score 1 if the model responds with ``don't know'', and 0 otherwise.

$F$ is the average of sufficiency and necessity. An $F$ score of 1.0 indicates that the model consistently relies on the cited supporting sentences to produce its answer, reflecting faithful reasoning rather than arbitrary guessing.

\subsubsection{Plausibility}

Plausibility, denoted as $P$, measures whether the model's supporting sentences yield a logically consistent and stable interpretation under benign transformations. We decompose it into two factors: \emph{paraphrase stability} and \emph{flip consistency}.

(i) \emph{Paraphrase Stability} ($PS$) measures whether the model preserves the same underlying spatial logic under lexical variability and linguistic difficulty. Similar to $F$, we use the supporting sentences generated by the model and create three variants using an LLM (GPT-5-mini~\citep{openai-gpt5}):   
(a) \textit{linguistically similar}, which paraphrases the sentence while retaining the original type of spatial expression, such as a clock-face direction
(e.g., the supporting sentence "T is above I at 2 o'clock" becomes "I is at the center and T is at the 2 o'clock position"); 
(b) \textit{Canonical}, which rewrites relations into simplified qualitative relations from our predefined relation set (e.g., "T upper-right position to I"); and 
(c) \textit{hinted}, which adds a hint to ease interpretation (e.g., hint: recognize the clockwise position here, 2 o'clock is usually upper-right from the center).
We run the model on these variants and compute $PS$ as the frequency of the most common answer among the three responses. This score lies in $[0,1]$, where higher values indicate more stable reasoning across paraphrased forms.

(ii) \emph{Flip Consistency} ($FC$) tests whether the model applies inverse spatial reasoning consistently with its original answer. Using an LLM, we generate a flipped question from the supporting sentences by reversing entity order for relation questions or negating the relation for yes/no questions. We compare the flipped answer with the expected inverse of the original response. $FC$ is 1 for a correct inverse, 0.5 for a partially correct answer in cases when the expected answer is multi-label, and 0 otherwise. Details are provided in Appendix~\ref{sec:implementation}.

$P$ is computed as the average of paraphrase stability and flip-consistency.
A $P$ score of 1.0 indicates that the model’s response remains stable under paraphrases and logically consistent under equivalent inversions.

\subsubsection{Complexity}

Complexity, denoted as $C$, measures the difficulty of a given spatial reasoning instance.
We quantify this difficulty for each spatial reasoning instance using a weighted combination of seven factors that capture the linguistic and multi-hop reasoning demands of the problem. Given the story, support sentences, and question, we use an LLM (GPT-5-mini~\citep{openai-gpt5}) to evaluate support sentences, identify entities, flag hard language patterns, and estimate coreference-based difficulty where applicable.
Based on these, we derive seven complementary factors. (i) Support Burden ($SB$) measures the number of supporting sentences derived by the model required for reasoning. (ii) Chain Length ($CL$) captures the actual multi-hop depth of the instance independent of the current model. This anchors complexity to the actual problem size. (iii) Selection Difficulty ($SD$) measures the fraction of story sentences that are not part of the supporting sentences, capturing the difficulty of filtering irrelevant context faced by the model. (iv) Hard Language ($HL$) captures the difficulty of linguistically complex spatial expressions, such as clock-face directions, directional phrases, nested relations, and coreference-dependent mentions. The LLM first identifies such expressions and assigns each one a difficulty score $d_i \in [0,1]$. Then we compute \(HL = 0.6 \cdot \max_i d_i + 0.4 \cdot \frac{1}{n}\sum_{i=1}^{n} d_i\), where $n$ is the number of identified expressions. 
% \[
% HL = 0.6 \cdot \max_i d_i + 0.4 \cdot \frac{1}{n}\sum_{i=1}^{n} d_i .
% \]
We give slightly greater weight to the maximum difficulty because a single challenging expression can become a reasoning bottleneck, while the mean captures cumulative linguistic difficulty across the story. We also analyze sensitivity to this weighting in Appendix~\ref{app:weight_sensitivity}.
(v) Diagonal Burden ($DB$) measures the fraction of extracted relations that are diagonal, i.e., composed of two simultaneous axis-aligned directions like lower-left.
(vi) Entity Load (EL) captures the burden of tracking multiple distinct entities. 
(vii) Coreference Difficulty ($CF$) estimates the burden of resolving entity references that require multi-step relational interpretation.
Since no instance-level complexity score is available, we estimate complexity using proxy signals. For frequency-based factors such as Support Burden, Chain Length, and Entity Load, we use the saturating function \(\mathrm{sat}(x,c)=\frac{x}{x+c}\) to normalize the frequency of the occurrences to a score in [0,1]. This normalization is needed because these factors are raw counts whose scale varies across datasets, and their difficulty contribution should increase sharply at lower values but saturate once the instance is already highly challenging. Here, \(c\) is a dataset-specific reference value set to the median of the corresponding feature so that moderate counts receive intermediate scores, while larger counts approach 1 without increasing linearly.
 These components are further combined into a normalized scalar $C \in [0,1]$, with dataset-specific weights reported in Appendix~\ref{sec:implementation}.
 We also analyze factor collinearity and error correlations in Appendix~\ref{sec:switching_signal}, finding mostly low-to-moderate collinearity and complementary trustworthiness-complexity behavior.

\noindent\textbf{Decision rule.} The switching policy avoids structured reasoning when an instance is simple and the model's natural language-based answer is reliable. We measure reliability as trustworthiness \(T = 0.6F + 0.4P\), where \(F\) denotes faithfulness and \(P\) denotes plausibility. Here, \(F\) receives greater weight because it directly measures whether the answer is supported by relevant evidence, while \(P\) captures the logical consistency of reasoning with that evidence. We analyze sensitivity to this weighting in Appendix~\ref{app:weight_sensitivity}. The policy retains natural language-based reasoning when complexity is low \((C < \tau_c)\) and trustworthiness is high \((T \geq \tau_t)\), and otherwise switches to grid-based reasoning. Thresholds \(\tau_t\) and \(\tau_c\) are tuned on the validation set for each dataset and model (Appendix~\ref{sec:implementation}).
For efficiency, we compute the signals using short-circuit evaluation. We first compute faithfulness and switch immediately if it is sufficient to determine \(T < \tau_t\). Otherwise, we compute complexity and switch if \(C \geq \tau_c\). Plausibility is computed only if neither signal has determined the decision, after which the final trustworthiness and complexity scores determine whether to retain text-based reasoning or switch to the grid. Further details are provided in Appendix~\ref{sec:short_circuit}.

\section{Experimental Results}

\noindent\textbf{Datasets:} We evaluate on three textual spatial reasoning benchmarks. \textbf{SpaRTUN} includes synthetic spatial narratives with Yes/No (YN) and Find Relations (FR) questions~\citep{spartun}. \textbf{StepGame} tests controlled multi-hop FR reasoning by requiring models
to select among nine spatial relations with increasing relational depth and
distractors~\citep{stepgameimproved}.
\textbf{ReSQ} contains YN questions over human-written real-world spatial descriptions~\citep{spartun}. Table~\ref{tab:dataset_stats} summarizes the statistics of the datasets. Due to cost, we evaluate randomly sampled subsets with a fixed seed of 0, preserving the original test-set distribution.

\noindent\textbf{Evaluation Metrics:} We report exact-match accuracy for all datasets. For multi-label FR questions, a prediction is correct only if it matches the complete ground-truth relation set~\citep{spartun}. We report accuracy per hop-depth for StepGame. We evaluate across different model families, including LLaMA3.1~\citep{grattafiori2024llama3herdmodels}, Qwen3~\citep{yang2025qwen3technicalreport}, and GPT-5.1~\citep{openai-gpt51}.

\begin{table}[t]
\centering
 \resizebox{0.99\columnwidth}{!}{
    \small
    \begin{tabular}{lccc}
    \toprule
    \textbf{Dataset} & \textbf{Eval. Samples} & \textbf{Full Test-Set} & \textbf{Max. Hop} \\
    \midrule
    SpaRTUN  & 1,121 & 5,551   & 7  \\
    StepGame & 5,000 & 100,000 & 10 \\
    ReSQ     & 610   & 610     & 2  \\
    \bottomrule
    \end{tabular}}
    \caption{Evaluation dataset statistics. Eval. Samples denotes the default
evaluation sample sizes; experiment-specific sample counts are reported in the
corresponding captions. Max. Hops denotes the maximum reasoning depth.}
    \label{tab:dataset_stats}
\end{table}

\begin{table*}[h!] \centering \resizebox{1.8\columnwidth}{!}{ \small \setlength{\tabcolsep}{5pt} \begin{tabular}{lccccccccccc} \toprule \textbf{Model / Setting} & \textbf{1} & \textbf{2} & \textbf{3} & \textbf{4} & \textbf{5} & \textbf{6} & \textbf{7} & \textbf{8} & \textbf{9} & \textbf{10} & \textbf{Overall} \\ \midrule LLaMA3.1-70B Text & \textbf{93} & 62 & 50 & 45 & 36 & 37 & 32 & 34 & 30 & 27 & 44.7 \\ LLaMA3.1-70B Relational Triples & 88 & 65 & 48 & 39 & 35 & 31 & 31 & 30 & 25 & 26 & 42.1 \\ LLaMA3.1-70B Coordinates Full & 85 & 79 & 80 & 76 & 74 & 68 & 71 & 68 & 69 & 67 & 73.7 \\ LLaMA3.1-70B Coordinates Pruned & 87 & 85 & \underline{84} & \underline{85} & \underline{86} & \underline{80} & \underline{81} & \underline{80} & \underline{78} & \underline{77} & \underline{82.3} \\ LLaMA3.1-70B Full Grid & 88 & \underline{87} & 80 & 73 & 76 & 70 & 77 & 74 & 73 & 70 & 76.7 \\ LLaMA3.1-70B Pruned Grid & \underline{92} & \textbf{92} & \textbf{88} & \textbf{87} & \textbf{87} & \textbf{83} & \textbf{88} & \textbf{84} & \textbf{83} & \textbf{82} & \textbf{86.7} \\ LLaMA3.1-70B ToT-CoT & 87 & 80 & 81 & 83 & 78 & 75 & 79 & 76 & 75 & 72 & 78.8 \\ \midrule Qwen3-32B Text & \underline{94.2} & 78.2 & 72.4 & 69.8 & 65.4 & 62.4 & 63.0 & 61.0 & 54.2 & 49.8 & 67.0 \\ Qwen3-32B Relational Triples & 89.2 & 77.4 & 68.6 & 63.4 & 58.2 & 53.6 & 56.2 & 51.6 & 42.8 & 45.2 & 60.6 \\ Qwen3-32B Coordinate Full & 87.6 & 80.4 & 77.6 & 71.8 & 77.8 & 72.4 & 74.4 & 71.0 & 69.6 & 73.0 & 75.6 \\ Qwen3-32B Coordinate Pruned & 85.8 & 83.6 & 82.8 & 81.2 & 84.4 & 80.4 & 82.4 & \underline{80.6} & 76.8 & 77.8 & 81.6 \\ Qwen3-32B Full Grid & 91.4 & \underline{87.8} & \textbf{87.8} & \textbf{84.6} & \textbf{86.0} & \textbf{82.4} & \textbf{84.0} & \textbf{82.0} & \textbf{80.6} & \textbf{79.8} & \textbf{84.6} \\ Qwen3-32B Pruned Grid & 91.8 & \textbf{90.2} & \underline{86.8} & \underline{83.6} & \underline{85.6} & \underline{81.8} & \underline{83.6} & 80.4 & \underline{78.4} & \underline{78.4} & \underline{84.1} \\ Qwen3-32B ToT-CoT & \textbf{94.6} & 83.0 & 75.2 & 66.4 & 66.2 & 62.4 & 62.4 & 56.6 & 59.2 & 54.2 & 68.0 \\ \midrule GPT-5.1 Text & \underline{99} & 92 & 85 & \underline{86} & 79 & 71 & 62 & 61 & 50 & 58 & 74.3 \\ GPT-5.1 Relational Triples & \textbf{100} & \underline{93} & 77 & 72 & 57 & 55 & 61 & 49 & 42 & 39 & 64.5 \\ GPT-5.1 Coordinates Full & 94 & 92 & 86 & \underline{86} & \textbf{88} & \textbf{89} & \underline{90} & \underline{80} & \underline{87} & 78 & 87.0 \\ GPT-5.1 Coordinates Pruned & 94 & 92 & 87 & \underline{86} & \textbf{88} & \textbf{89} & \underline{90} & \underline{80} & \underline{87} & 79 & 87.2 \\ GPT-5.1 Full Grid & 92 & 92 & \underline{88} & 85 & 86 & 86 & \textbf{92} & \textbf{85} & \textbf{89} & \underline{83} & \underline{87.8} \\ GPT-5.1 Pruned Grid & 92 & 92 & \underline{88} & 85 & 86 & \underline{88} & \textbf{92} & \textbf{85} & \textbf{89} & \textbf{84} & \textbf{88.1} \\ GPT-5.1 ToT-CoT & 98 & \textbf{95} & \textbf{92} & \textbf{91} & \underline{87} & \textbf{89} & 86 & 77 & 75 & 70 & 86.0 \\ \bottomrule \end{tabular}} \caption{Accuracy (\%) across (k)-hop levels on StepGame for text-, triple-, coordinate-, and grid-based representations, compared with the Tree-of-Thought Chain-of-Thought (ToT-CoT) prompting method~\citep{stepgameimproved}. For all non-GPT-5.1 models, we evaluate (n=5{,}000) examples, with 500 examples per hop level. For GPT-5.1, we evaluate (n=1{,}250) examples for cost reasons, with 125 examples per hop level. Overall denotes the mean accuracy across (k=1)--(10). The best result is shown in bold and the second-best result is underlined within each model block.} \label{tab:accuracy_khop_stepgame_main} \end{table*}

\textbf{RQ1. Can mapping to grids simplify multi-hop reasoning and improve performance?}
\label{sec:grid_benefits} 
Table~\ref{tab:accuracy_khop_stepgame_main} demonstrates the results on StepGame. We observe that the grid-based representations yield the strongest overall performance, especially at higher hops, where text-only reasoning declines sharply. 
\textsc{LLaMA3.1-70B} benefits most from pruned grids, suggesting that reducing irrelevant spatial context helps non-reasoning models. Representations requiring more numerical manipulation, such as coordinate layouts, are more effective for stronger models such as \textsc{Qwen3-32B} and \textsc{GPT-5.1}~\citep{yang2025qwen3technicalreport,openai-gpt51}. Appendix~\ref{app:small_lm_modalities} further shows that this pattern holds for small language models, where grid-based representations improve overall accuracy by up to 25\%.

Table~\ref{tab:spartun_resq_results} further demonstrates that structured spatial representations improve reasoning on SpaRTUN, especially for Find Relations (FR) questions, where models must recover multiple object relations. Pruned grids also perform well on Yes/No (YN) questions, often achieving the best or second-best accuracy for each model. Reasoning models such as \textsc{Qwen3-32B} benefit substantially from grid-based reasoning, improving overall accuracy by up to 8\%, due to more accurate relation extraction during grid construction. For GPT-5.1, text-only reasoning is strong, but pruned grids still improve performance by focusing on the most relevant spatial objects. Finally, Chain of Symbol (CoS) is also competitive with grid-based reasoning in several settings, suggesting that symbolic intermediate forms can help when questions require tracking multiple relations. Appendix~\ref{app:small_lm_modalities} shows that for smaller models, grid-based reasoning remains competitive with text-based reasoning on SpaRTUN. We also analyze representation quality and grid construction behavior in Appendix~\ref{app:ablation}. 

\begin{table}[t]
\vspace{-6pt}
\centering
\tiny
\setlength{\tabcolsep}{1.6pt}
\renewcommand{\arraystretch}{0.94}
\resizebox{0.9\columnwidth}{!}{%
\begin{tabular}{lccccc}
\toprule
& \multicolumn{3}{c}{\textbf{SpaRTUN}} & \multicolumn{2}{c}{\textbf{ReSQ}} \\
\cmidrule(lr){2-4}\cmidrule(lr){5-6}
\textbf{Model / Setting} & \textbf{YN} & \textbf{FR} & \textbf{Overall} & \textbf{GPT-grid } & \textbf{SM-grid} \\
\midrule
LLaMA3.1-70B Text & \underline{75.45} & 32.50 & 55.76 & 77.05 & 77.05 \\
LLaMA3.1-70B Rel. Trip. & 74.80 & 32.30 & 55.32 & 76.23 & 69.84 \\
LLaMA3.1-70B CoS & 75.41 & 40.31 & 59.59 & -- & -- \\
LLaMA3.1-70B FG & \textbf{76.27} & \textbf{40.98} & \textbf{60.21} & \textbf{80.33} & \textbf{78.20} \\
LLaMA3.1-70B PG & \textbf{76.27} & \underline{40.78} & \underline{60.12} & \underline{80.00} & \textbf{78.20} \\

\midrule
Qwen3-32B Text & \textbf{85.30} & 48.80 & 68.85 & 80.30 & \textbf{80.30} \\
Qwen3-32B Rel. Trip. & 77.40 & 62.40 & 70.67 & 68.85 & 68.85 \\
Qwen3-32B CoS & 78.60 & 43.90 & 62.80 & -- & -- \\
Qwen3-32B FG & 78.80 & \underline{69.80} & \underline{74.79} & \underline{83.28} & 79.50 \\
Qwen3-32B PG & \underline{82.40} & \textbf{70.40} & \textbf{77.07} & \textbf{85.25} & \underline{79.83} \\
\midrule
GPT-5.1 Text & 75.35 & \textbf{63.68} & 70.00 & 82.60 & -- \\
GPT-5.1 Rel. Trip. & 79.08 & 55.21 & 68.14 & 79.18 & -- \\
GPT-5.1 CoS & \underline{81.14} & 59.43 & \underline{71.19} & -- & -- \\
GPT-5.1 FG & 77.45 & 55.98 & 67.61 & \textbf{85.41} & -- \\
GPT-5.1 PG & \textbf{81.37} & \underline{60.23} & \textbf{71.68} & \underline{84.59} & -- \\
\bottomrule
\end{tabular}%
}
\caption{Accuracy (\%) on SpaRTUN and ReSQ. SpaRTUN reports Yes/No (YN), Find Relations (FR), and overall accuracy; due to cost, we evaluate distribution-preserving subsets with \(n=565\) questions for GPT-5.1 and \(n=1{,}121\) for all other models. For ReSQ, we use the full test set. GPT-grid denotes GPT-5.1 grid construction, SM-grid denotes same-model grid construction, FG denotes full-grid, PG denotes pruned-grid, and Rel. Trip. denotes relational triples. Best and second-best results are made bold and underlined.}
\label{tab:spartun_resq_results}
\end{table}

We evaluate ReSQ to test generalizability on realistic spatial descriptions and report results in Table~\ref{tab:spartun_resq_results}. Here, grids are constructed directly from extracted relations using LLMs. 
GPT-based construction performs best, while same-model construction remains competitive with the baselines, showing that open-source models can be used for grid construction with minimal accuracy loss. 
Since ReSQ contains only up to two hops of reasoning, the accuracy gains are smaller than those on StepGame. However, grid-based representations consistently outperform the baselines by up to 5\%. Appendix~\ref{app:small_lm_modalities} shows similar gains with grid-based reasoning on ReSQ for small language models.
Overall, grids help most when reasoning requires composing many relations, while pruning reduces noisy context. Text-only reasoning often fails on multi-step composition, while grid-based reasoning can fail due to relation extraction or grid construction errors. Because the two modalities fail for different reasons, switching between them can be useful. We further evaluate the transformation pipeline by separately examining relation extraction and grid construction on StepGame and SpaRTUN, for which ground-truth relation triples are available. 

Relation-extraction F1 ranges from 94.3 to 95.0 on StepGame and from 77.5 to 87.5 on SpaRTUN across models. To isolate grid construction, we also construct grids from ground-truth relations, using 2D grids for StepGame and grids with symbolically encoded 3D spatial relations for SpaRTUN. Grid-construction F1 is 100.00 on StepGame and 95.16 on SpaRTUN. When constructed from extracted relations, the F1 scores for construction range from 86.68 to 89.68 on StepGame and from 72.40 to 78.39 on SpaRTUN. These are the grids used for grid-based reasoning. Further details are in Appendix~\ref{app:ablation}.

\begin{figure*}[t]
    \centering
    \includegraphics[width=0.7\textwidth]{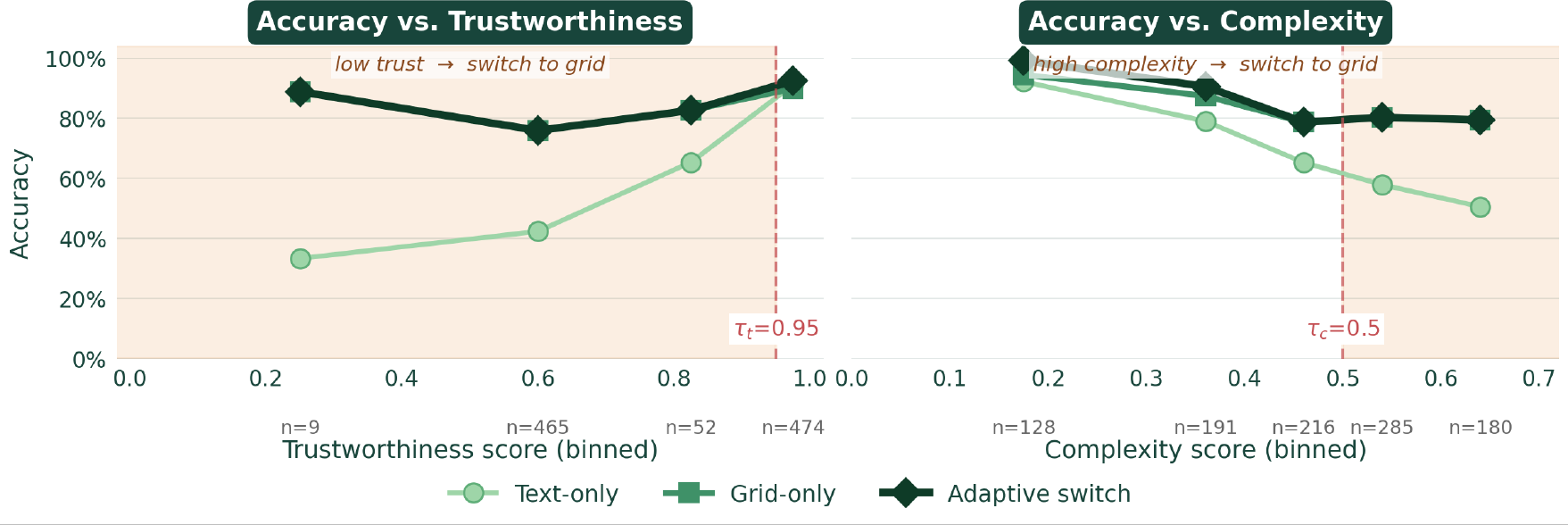}
    \vspace{-6pt}
    \caption{
    \textbf{Trustworthiness and complexity predict when switching helps.}
    Accuracy of \textsc{Qwen3-32B} on \textsc{StepGame} ($n{=}1{,}000$) across binned trustworthiness scores (left) and complexity scores (right). The $n$ labels on the x-axis denote the number of instances in each bin. Red lines show the switching thresholds, $\tau_t{=}0.95$ and $\tau_c{=}0.50$, and peach regions mark where the policy switches to grid-based reasoning.
    }
    \label{fig:trust_complexity_motivation}

\end{figure*}

\noindent\textbf{RQ2\@. Is the switching policy based on trustworthiness and complexity effective?}
\label{sec:switching} 
Table~\ref{tab:combined_reasoning_results} demonstrates that across models, adaptive switching often matches or outperforms grid-only reasoning while switching to the grid modality for only a subset of instances. The oracle result provides an upper bound by selecting the correct route whenever either text-only or grid-based reasoning gives the correct answer. This shows that larger gains are possible if switching decisions can more accurately identify when each modality is useful.
Figure~\ref{fig:trust_complexity_motivation} further explains why trustworthiness and complexity signals help. Text-only accuracy declines in low-trust and high-complexity regions, while adaptive switching remains more stable by routing difficult cases to the grid modality. Additional analysis in Appendix~\ref{sec:trust_complexity_alignment},~\ref{app:factor_correlations} and~\ref{app:error_taxonomy} show that the two signals are complementary: trustworthiness and complexity individually result in weaker routing accuracy in most settings, while the combined adaptive policy gives the strongest performance across models. Their correlations with text-only accuracy also follow the expected pattern, with trust-related factors generally positive and complexity-related factors generally negative, indicating that the policy captures both unreliable answers and intrinsically difficult instances.

\begin{table}[t]
\vspace{-8pt}
\centering
\footnotesize
\setlength{\tabcolsep}{2.2pt}
\renewcommand{\arraystretch}{0.92}
\resizebox{\columnwidth}{!}{%
\begin{tabular}{lcccc|cccc}
\toprule
& \multicolumn{4}{c|}{\textbf{Acc.}} & \multicolumn{4}{c}{\textbf{Rate}} \\
\cmidrule(lr){2-5}\cmidrule(lr){6-9}
\textbf{Setting} & \textbf{Txt} & \textbf{G} & \textbf{AS} & \textbf{Orc} & \textbf{S} & \textbf{T} & \textbf{C} & \textbf{B} \\
\midrule
SG LLaMA3.1-70B & 42.4 & 82.8 & 83.6 & 86.8 & 78.8 & 40.1 & 10.7 & 49.2 \\
SG Qwen3-32B    & 63.6 & 84.0 & 84.4 & 91.2 & 69.6 & 58.6 & 15.5 & 25.9 \\ 
SG GPT-5.1      & 71.2 & 83.6 & 85.2 & 91.6 & 80.4 & 8.0  & 41.3 & 50.7 \\ 
\midrule
ST LLaMA3.1-70B & 56.4 & 60.1 & 60.8 & 76.2 & 72.3 & 96.9 & 0.6  & 2.5 \\
ST Qwen3-32B    & 67.5 & 74.5 & 74.5 & 88.6 & 55.5 & 53.0 & 25.2 & 21.9 \\
ST GPT-5.1      & 72.9 & 76.5 & 77.4 & 91.0 & 69.9 & 66.4 & 8.6  & 25.0 \\
\bottomrule
\end{tabular}%
}
\caption{Accuracy and switching behavior on \textsc{StepGame} (SG) and
\textsc{SpaRTUN} (ST), where $n$ denotes the number of test instances.
Switching results are reported on \textsc{StepGame} with $n{=}1{,}000$ for
\textsc{LLaMA3.1-70B} and \textsc{Qwen3-32B}, and $n{=}500$ for \textsc{GPT-5.1} due to cost. For
\textsc{SpaRTUN}, we use $n{=}891$ for \textsc{LLaMA3.1-70B} and \textsc{Qwen3-32B}, and $n{=}561$
for GPT-5.1, while preserving the original test-set distribution. Txt, G, AS,
and Orc denote text-only, grid-only, adaptive-switching, and oracle accuracy,
respectively. S denotes the overall switch rate; T, C, and B denote the
percentages of switched instances attributed to low trustworthiness, high
complexity, and both signals, respectively.}
\label{tab:combined_reasoning_results}
\vspace{-10pt}
\end{table}

\noindent\textbf{Failure Modes.} To better understand when switching helps, we analyze failures based on the original problem and model reasoning. 
Switching is most effective when a model fails to reason correctly in text on a multi-hop reasoning problem. In such cases, our switching mechanism recovers up to 83.6\% of switched multi-hop reasoning failures. By contrast, failures involving linguistically difficult cases remain harder to recover because many residual errors arise during relation extraction in grid construction. A detailed discussion can be found in Appendix~\ref{app:error_taxonomy}.
Overall, these findings show that trustworthiness and complexity provide useful signals for modality switching.

\noindent\textbf{RQ3. How can adaptive switching balance accuracy and computational cost?}
Adaptive switching selects the more expensive grid-based reasoning route for instances where text-only reasoning is unreliable or the problem is complex. Switching can improve accuracy but incurs additional signal-computation costs, which are reduced through the short-circuit evaluation described in Section~\ref{sec: decision}. The resulting cost--accuracy balance therefore depends on how often the grid route is used and which signals are required to trigger it.
This trade-off varies across datasets. On SpaRTUN, adaptive switching, including switching costs, reduces average token cost by 6.4--12.0\% relative to grid-only reasoning while matching or improving accuracy by up to 0.9\%. On StepGame, token cost ranges from a 0.4\% reduction to a 10.6\% increase, with accuracy gains of 0.4--1.6\%. In particular, \textsc{GPT-5.1} achieves the largest StepGame accuracy gain (+1.6\%) but also the highest token overhead (+10.6\%), partly because more of its switches require the relatively costly complexity signal. Full token-cost breakdowns for trust-only, complexity-only, and combined switching are provided in Appendix~\ref{sec:rq3_cost_analysis}.
Overall, adaptive switching is not always cheaper than grid-only reasoning; instead, it provides an accuracy--cost allocation mechanism that avoids structured reasoning when unnecessary and is most cost-effective when the structured route is substantially more expensive.

\noindent\textbf{RQ4. Do the switching signals align with human judgments?} We test whether the complexity, trustworthiness, faithfulness, and
plausibility signals used for switching align with human judgments of the
same properties. For this evaluation, we sample $n{=}240$ instances from StepGame
and SpaRTUN across \textsc{Qwen3-32B} and \textsc{LLaMA3.1-70B}, with 60
instances for each model and dataset combination. Each instance includes the
story, question, model answer, and model-selected support sentences.
Three annotators first rate complexity based on the story and question alone. Next, they rate faithfulness, plausibility, and trustworthiness based on the model answer and its supporting sentences. Each dimension is rated from 1 to 5, where higher ratings indicate greater
complexity, faithfulness, plausibility, or trustworthiness. 

Since the switching signals and human ratings use different
scales, we measure rank-based alignment with Spearman's $\rho$, using the
mean human rating per instance. We use Krippendorff's $\alpha$ to measure
inter-annotator agreement. Complete instructions and rating criteria are
provided in Appendix~\ref{app:human_evaluation}.
\begin{table}[t]
\centering
\small
\renewcommand{\arraystretch}{1.05}
\resizebox{\columnwidth}{!}{%
\begin{tabular}{llcccc}
\toprule
\textbf{Model / Agreement} & \textbf{Dataset} &
\textbf{C} &
\textbf{T} &
\textbf{F} &
\textbf{P} \\
\midrule
Qwen3-32B & StepGame & 0.63 & \textbf{0.69} & \textbf{0.66} & 0.49 \\
Qwen3-32B & SpaRTUN & \textbf{0.69} & 0.62 & 0.57 & \textbf{0.58} \\
LLaMA3.1-70B & StepGame & 0.61 & 0.64 & 0.52 & 0.48 \\
LLaMA3.1-70B & SpaRTUN & 0.62 & 0.52 & 0.58 & 0.36 \\
\midrule
Krippendorff's $\alpha$ & Agreement &
\textbf{0.67} & 0.65 & 0.61 & 0.56 \\
\bottomrule
\end{tabular}%
}
\caption{Human evaluation of the proposed switching signals across model--dataset settings ($n{=}60$ per setting). C, T, F, and P denote complexity, trustworthiness, faithfulness, and plausibility, respectively; Agreement denotes inter-annotator agreement.}
\label{tab:human_eval}
\end{table}
We observe positive correlations between the switching signals and human
judgments across all settings, as shown in Table~\ref{tab:human_eval}, with
Spearman's $\rho$ ranging from 0.36 to 0.69. The strongest
alignment occurs for Qwen3-32B trustworthiness on StepGame and complexity on
SpaRTUN ($\rho=0.69$). Plausibility has the weakest alignment
($\rho=0.36$) and agreement ($\alpha=0.56$), likely because annotators judge
the logical consistency of reasoning from a single response, while the
switching signal also evaluates the consistency of model answers across
controlled variations. Complexity has the highest agreement ($\alpha=0.67$), indicating greater consistency among annotators.

Overall, we view human evaluation as complementary validation of the switching
signals. It assesses whether the signals correspond to human perceptions of
the same properties, while the downstream switching results directly test
whether they lead to useful routing decisions. Together, these evaluations
support using the signals to determine when a different reasoning
representation may be beneficial, rather than interpreting them as global
measures of model trustworthiness.

\section{Conclusion}

We study whether grounding spatial narratives into geometry-preserving
representations improves language model reasoning. Across
\textsc{StepGame}, \textsc{SpaRTUN}, and \textsc{ReSQ}, grid-based grounding
consistently improves over natural language-based reasoning, with larger gains
as reasoning depth and spatial complexity increase. Our findings
show that geometry-preserving representations provide an effective medium for
spatial reasoning. We further investigate when such grounding is beneficial
and develop an adaptive switching policy based on trustworthiness and
complexity signals. Our findings show that these signals effectively guide
modality selection and avoid unnecessary switches to the more costly grid
representation. Therefore, adaptive modality selection can lead to more
reliable and efficient spatial reasoning.

\section*{Limitations}

Our framework depends on reliable intermediate spatial structure construction. Grid-based reasoning is helpful only when relation extraction and grid construction accurately reflect the original narrative; otherwise, extraction errors can propagate and reduce downstream accuracy. We partially mitigate this through line-by-line relation extraction, verification, and coreference-aware processing, but future work could improve this component through spatial-relation extraction fine-tuning.

Another limitation is that adaptive switching introduces computational overhead because trustworthiness and complexity signals must be computed before routing. Although short-circuit computation reduces unnecessary checks, the switching signal is not free and can sometimes offset the savings from avoiding grid construction. Future work could develop cheaper routing signals or learned routing policies that better predict when structured reasoning will improve the answer.
\section*{Acknowledgements}
This project is partially supported by the Office of Naval Research (ONR) under grant N00014-23-1-2417, the Michigan State University Distinguished Fellowship, and Lambda Labs. Any opinions, findings, conclusions, or recommendations expressed in this material are those of the authors and do not necessarily reflect the views of the Office of Naval Research, Michigan State University, or Lambda Labs.

\bibliography{referece}

\appendix

\appendix
\section{Experimental Setup}
\label{sec:experiments}

All local experiments were conducted on a single NVIDIA GH200 GPU with 96GB memory. Unless otherwise stated, local inference used greedy decoding with temperature $0.0$. GPT-5.1~\citep{openai-gpt51} was accessed through the OpenAI Responses API with reasoning effort set to \texttt{none}. The switching thresholds are selected on the validation split and then fixed for test evaluation. To ensure reproducibility, all sampled subsets use a fixed random seed of 0, and the same evaluated instances are used across comparable settings.
We use the Qwen3 family ~\citep{yang2025qwen3technicalreport}, including Qwen3-8B, Qwen3-14B, and Qwen3-32B, with temperature 0.7 to enable their explicit reasoning mode and LLaMA3.1~\citep{grattafiori2024llama3herdmodels}, including LLaMA3.1-8B and LLaMA3.1-70B, with temperature 0.0.
All prompts and evaluation scripts were identical across experimental conditions to ensure 
fair comparison between reasoning modalities; these are available in the code for reproducibility.

\section{Switching Implementation}
\label{sec:implementation}
This section details how the adaptive switching policy is implemented, including the construction of the trustworthiness and complexity signals, the dataset-specific weighting of complexity factors, and the threshold-selection procedure. We also describe the short-circuit computation used to reduce switching overhead and analyze how the signals align with text-only accuracy.

\textbf{Trust Signal}
We define trustworthiness as \(T = 0.6F + 0.4P\), giving slightly higher weight to faithfulness than plausibility. This choice reflects the role of the two signals in the switching decision. Faithfulness directly checks whether the model's answer is grounded in the story evidence, so it should dominate the trust score: if the answer is not supported by the relevant spatial facts, then a high plausibility or stability score may only indicate that the model is consistently confident in an unsupported answer. Plausibility is still important because it tests whether the answer remains stable under benign transformations, but we treat it as a secondary signal that can reinforce, rather than override, evidence grounding. Empirically, we use a conservative gap between the two weights: changing either weight by \(0.1\) would make the score closer to an equal weighting, while a larger gap would make plausibility too weak to affect borderline cases. Thus, the \(0.6/0.4\) split encodes a minimal preference for faithfulness while still allowing plausibility to influence the final trust decision.

\textbf{Dataset-specific $C$ configuration.}
The components of $C$ are selectively activated and weighted based on the
dominant difficulty sources in each dataset. Table~\ref{tab:c3_weights}
reports the weight configuration used in our runs. For StepGame, the weights
emphasize structural multi-hop difficulty: support burden (SB), selection
difficulty (SD), chain length (CL), hard language (HL), and diagonal burden
(DB) are active. In particular, SD captures the difficulty of identifying the
relevant supporting chain among distractor relations. Entity load (EL) is not
used because each StepGame relation sentence typically involves two entities,
and entity-related difficulty is already largely captured by chain length,
support burden, and selection difficulty, making EL redundant in this setting.

For SpaRTUN, the weights emphasize linguistic and grounding difficulty.
SpaRTUN contains richer spatial language, logical quantifiers, mixed
topological and directional relations, and entity references that can require
coreference-style interpretation. Accordingly, hard language (HL) and
coreference difficulty (CF) receive the largest weights, while chain length
(CL) and entity load (EL) provide complementary structural information.
Selection difficulty (SD) is not included because it is less independently
expressed in SpaRTUN and is largely redundant with the other factors, while
the dominant difficulty lies more in linguistic interpretation and multi-step
reasoning. Support burden (SB) and diagonal burden (DB) are also not used.

\begin{table}[h]
\centering
\small
\setlength{\tabcolsep}{1mm}
\begin{tabular}{lccccccc}
    \toprule
    \textbf{Dataset} & \textbf{SB} & \textbf{SD} & \textbf{CL} & \textbf{HL} & \textbf{DB} & \textbf{EL} & \textbf{CF} \\
    \midrule
    StepGame & 0.22 & 0.17 & 0.22 & 0.28 & 0.11 & 0.00 & 0.00 \\
    SpaRTUN  & 0.00 & 0.00 & 0.12 & 0.37 & 0.00 & 0.12 & 0.37 \\
    \bottomrule
\end{tabular}
\caption{$C$ weight configurations per dataset. SB = Support Burden, SD = Selection Difficulty, CL = Chain Length, HL = Hard Language, DB = Diagonal Burden, EL = Entity Load, and CF = Coreference Difficulty.}
\label{tab:c3_weights}
\end{table}

Diagonal Burden (DB) is not considered for SpaRTUN because its relation
inventory is based on RCC8 and directional primitives rather than explicit
diagonal labels such as lower-left or upper-right. Coreference Difficulty
(CF) is not considered for StepGame because entities are explicitly named and
do not introduce reference ambiguity. For extensions to new datasets, factors
can first be screened using collinearity analysis to remove redundant signals
and then retained according to the dominant sources of difficulty in the
dataset.

 \subsection{Sensitivity to Weight Choices}
\label{app:weight_sensitivity}

The weights used in the trustworthiness and hard-language scores are
conceptually motivated rather than optimized hyperparameters. For
trustworthiness, faithfulness receives slightly greater weight because
identifying the relevant supporting evidence is a prerequisite for reliable
reasoning, while plausibility provides a complementary consistency check. For
hard language, the maximum expression-level difficulty receives slightly
greater weight because a single difficult spatial expression can become a
reasoning bottleneck, while the mean captures cumulative linguistic difficulty.

To evaluate sensitivity to these choices, we vary the first weight from
$0$ to $1$ in increments of $0.1$, with the second weight set to its
complement, and rerun adaptive switching. We use validation sets of $n{=}350$
StepGame and $n{=}345$ SpaRTUN instances following the distributions used in our
experiments.

\paragraph{Faithfulness--plausibility weights.}
Table~\ref{tab:fp_weight_sensitivity} compares the proposed $(0.6,0.4)$
weighting against the best validation configuration.

\begin{table}[h]
\centering
\small
\setlength{\tabcolsep}{2.5pt}
\resizebox{\columnwidth}{!}{%
\begin{tabular}{llccccc}
\toprule
\textbf{Dataset} & \textbf{Model} &
\textbf{Proposed} &
\textbf{Best} &
\textbf{Prop. Acc.} &
\textbf{Best Acc.} &
\textbf{Diff.} \\
\midrule
StepGame & Qwen3-32B     & (0.6, 0.4) & (1.0, 0.0) & 83.6 & 83.6 & +0.0 \\
StepGame & LLaMA3.1-70B  & (0.6, 0.4) & (1.0, 0.0) & 82.0 & 82.4 & +0.4 \\
StepGame & GPT-5.1       & (0.6, 0.4) & (0.6, 0.4) & 82.8 & 82.8 & +0.0 \\
SpaRTUN  & Qwen3-32B     & (0.6, 0.4) & (0.5, 0.5) & 72.6 & 73.0 & +0.4 \\
SpaRTUN  & LLaMA3.1-70B  & (0.6, 0.4) & (0.7, 0.3) & 58.7 & 59.1 & +0.4 \\
SpaRTUN  & GPT-5.1       & (0.6, 0.4) & (0.6, 0.4) & 76.1 & 76.1 & +0.0 \\
\bottomrule
\end{tabular}%
}
\caption{Sensitivity of adaptive-switching accuracy to the
faithfulness--plausibility weights. Proposed denotes the $(0.6,0.4)$ setting
used in the main experiments, while Best reports the highest-performing
validation configuration.}
\label{tab:fp_weight_sensitivity}
\end{table}

The proposed $(0.6,0.4)$ setting is optimal or within 0.4\% of the best
configuration in every setting. This suggests that routing decisions are
relatively insensitive to moderate changes in the relative contributions of
faithfulness and plausibility.

\paragraph{Hard-language weights.}
We similarly vary the relative contribution of the maximum and mean
expression-level difficulty used to compute the hard-language component.
Results are shown in Table~\ref{tab:hl_weight_sensitivity}.

\begin{table}[h]
\centering
\small
\setlength{\tabcolsep}{2.5pt}
\resizebox{\columnwidth}{!}{%
\begin{tabular}{llccccc}
\toprule
\textbf{Dataset} & \textbf{Model} &
\textbf{Proposed} &
\textbf{Best} &
\textbf{Prop. Acc.} &
\textbf{Best Acc.} &
\textbf{Diff.} \\
\midrule
StepGame & Qwen3-32B     & (0.6, 0.4) & (0.5, 0.5)$^\dagger$ & 83.6 & 84.0 & +0.4 \\
StepGame & LLaMA3.1-70B  & (0.6, 0.4) & (0.5, 0.5)$^\dagger$ & 82.0 & 82.0 & +0.0 \\
StepGame & GPT-5.1       & (0.6, 0.4) & (0.6, 0.4)$^\dagger$ & 82.8 & 82.8 & +0.0 \\
SpaRTUN  & Qwen3-32B     & (0.6, 0.4) & (0.7, 0.3)           & 72.6 & 73.5 & +0.9 \\
SpaRTUN  & LLaMA3.1-70B  & (0.6, 0.4) & (0.6, 0.4)           & 58.7 & 58.7 & +0.0 \\
SpaRTUN  & GPT-5.1       & (0.6, 0.4) & (0.5, 0.5)           & 76.1 & 76.5 & +0.4 \\
\bottomrule
\end{tabular}%
}
\caption{Sensitivity of adaptive-switching accuracy to the hard-language
weights. $^\dagger$Several neighboring configurations produce the same or
nearly identical routing outcomes; one representative best configuration is
reported.}
\label{tab:hl_weight_sensitivity}
\end{table}

Most hard-language weight configurations produce similar routing decisions.
The proposed setting remains within 0.9\% of the best configuration across all
settings. StepGame shows relatively little sensitivity because linguistic
difficulty is limited and often overlaps with structural multi-hop difficulty,
whereas SpaRTUN contains more heterogeneous linguistic expressions and shows
small model-specific differences. Overall, both $(0.6,0.4)$ choices provide
robust defaults rather than heavily tuned hyperparameters.

\subsection{Sensitivity to the Saturation Function}
\label{app:saturation_sensitivity}

For non-negative count-based complexity factors, we use
$\mathrm{sat}(x,c)=x/(x+c)$ to model diminishing increases in difficulty.
The intuition is that increasing a short reasoning chain up to a moderate
length can substantially increase difficulty, whereas further increasing an
already long chain should have a smaller marginal effect. The function is
bounded in $[0,1)$ and allows a feature-specific half-saturation point $c$.

To test whether our results depend strongly on this functional form, we rerun
adaptive switching using sigmoid and logarithmic alternatives on the same test
splits used in our main switching experiments. Table~\ref{tab:saturation_ablation}
reports adaptive accuracy and the percentage of examples switched to grid
reasoning.

\begin{table}[h]
\centering
\small
\setlength{\tabcolsep}{2.5pt}
\resizebox{\columnwidth}{!}{%
\begin{tabular}{llcccc}
\toprule
\textbf{Dataset} & \textbf{Model} &
\textbf{Grid} &
\textbf{$x/(x+c)$} &
\textbf{Sigmoid} &
\textbf{Logarithmic} \\
\midrule
StepGame & Qwen3-32B
& 84.0 & \textbf{84.4 / 69.6} & 84.5 / 74.4 & 83.6 / 88.0 \\

StepGame & LLaMA3.1-70B
& 82.8 & 83.6 / 78.8 & \textbf{84.2 / 78.2} & 83.8 / 83.9 \\

SpaRTUN & Qwen3-32B
& 74.5 & \textbf{74.5 / 55.5} & 73.4 / 49.7 & 74.2 / 55.9 \\

SpaRTUN & LLaMA3.1-70B
& 60.1 & \textbf{60.8 / 72.3} & 61.3 / 79.6 & 61.6 / 75.8 \\
\bottomrule
\end{tabular}%
}
\caption{Sensitivity to the normalization function used for count-based
complexity factors. Adaptive configurations are reported as
accuracy / switch rate (\%). Grid denotes grid-only reasoning. Bold denotes
configurations that match or exceed grid-only accuracy while requiring fewer
switches.}
\label{tab:saturation_ablation}
\end{table}

Accuracy varies by only 0.6--1.1\% across normalization functions, although
the switching rate varies more substantially. The proposed $x/(x+c)$
normalization lies on the accuracy--switching trade-off frontier in three of
the four settings. We therefore use it as a simple and interpretable default,
rather than assuming it to be uniquely optimal. A sigmoid provides similar
saturation but introduces an additional slope parameter, while a logarithmic
transformation requires an additional normalization choice because it is
unbounded.

\textbf{Saturating function caps.}
For frequency-based candidate factors, we use
$\text{sat}(x,c)=x/(x+c)$, where $c$ is the half-saturation point.
For SB and CL on StepGame, we set $c=5$, as difficulty increases sharply
up to 5 hops and exhibits diminishing returns beyond that. For SpaRTUN,
where SB is defined using the number of identified support sentences rather
than $k_{\text{hop}}$, we set $c=3$ to reflect shorter support chains.
For EL, we set $c=6$ for StepGame and $c=10$ for SpaRTUN, aligned with
the median entity count in each dataset. These definitions are provided for
the full set of candidate complexity factors; factors assigned zero weight
in Table~\ref{tab:c3_weights} do not contribute to the final dataset-specific
complexity score.

\textbf{Threshold selection.}
For each instance, the switching rule keeps natural language-based reasoning only when the text answer is sufficiently trustworthy and the instance is not too complex, i.e., \(T \geq \tau_t\) and \(C < \tau_c\); otherwise, it switches to grid-based reasoning. We tune \((\tau_t,\tau_c)\) on a held-out validation split and report results on a disjoint test split. The validation split is stratified by question type and hop count where applicable, so the threshold search preserves the dataset distribution.
We perform a grid search over \(\tau_t,\tau_c \in \{0.05,0.10,\ldots,0.95\}\). For each candidate pair, we compute adaptive accuracy, switch rate, total routing cost, and the distribution of switches attributed to trustworthiness, complexity, or both. Total routing cost includes the cost of the switching signals together with the selected reasoning route, with the grid pipeline incurred only for switched instances.
We select thresholds using the following criteria. First, we retain candidates whose adaptive validation accuracy matches or exceeds the grid-only validation accuracy; if none satisfy this condition, we retain the highest-accuracy candidates. Among these candidates, we next prefer settings whose resulting switch rate keeps the total adaptive routing cost at or below the cost of always using the grid. If no accuracy-preserving candidate satisfies this cost condition, we retain the lowest-cost candidates among the accurate settings rather than sacrificing accuracy solely to satisfy the cost constraint.
We then consider the balance of the switching signals. We avoid settings in which one signal completely dominates the routing decision: candidates with a switch source contributing \(0\%\), or with a single source accounting for more than \(95\%\) of switches, are excluded when alternatives are available. This serves as a secondary constraint to ensure that both trustworthiness and complexity contribute meaningfully to the policy. Among the remaining candidates, we select the highest-accuracy setting, breaking further ties by lower total routing cost and then lower switch rate.
The thresholds used for the main model settings are reported together with the routing-cost analysis in Table~\ref{tab:routing_costs}. In additional smaller-model experiments, we use \((\tau_t,\tau_c)=(0.85,0.45)\) for Qwen3-8B and Qwen3-14B, and \((0.75,0.30)\) for LLaMA3.1-8B. The lower complexity threshold for LLaMA3.1-8B reflects its lower text-only robustness on moderately complex instances. Across exploratory runs, we observed that \((0.85,0.50)\) provides a reasonable default across datasets and models, and we suggest using it as the starting point for new datasets or extended experiments.
 
% \textbf{Pipeline and short-circuit.}
% The switching decision is computed in three stages, with parallel LLM calls within each stage. In Stage 1, the model produces an initial answer and identifies support sentences S. In Stage 2, using S, four calls run in parallel: the sufficiency check, the necessity check, GPT-5-mini generating paraphrase variants and a flip question, and GPT-5-mini scoring hard-language spans. In Stage 3, the paraphrase variants and flip question are answered in parallel to compute P, while F and C are already available. 
\begin{table*}[t] \centering \small \setlength{\tabcolsep}{4pt} \renewcommand{\arraystretch}{0.95} \begin{tabular}{llccc} \toprule \textbf{Dataset} & \textbf{Factor} & \textbf{Qwen3-32B} & \textbf{LLaMA3.1-70B} & \textbf{GPT-5.1} \\ \midrule StepGame & \textit{Text acc.} & 63.6 & 42.4 & 71.2 \\ \cmidrule(lr){2-5} StepGame & \textbf{Trustworthiness} & $+0.553$ & $+0.313$ & $+0.533$ \\ StepGame & \quad Faithfulness \(F\) & $+0.560$ & $+0.329$ & $+0.568$ \\ StepGame & \quad \quad Sufficiency \(S\) & $+0.589$ & $+0.348$ & $+0.568$ \\ StepGame & \quad \quad Necessity \(N\) & $+0.031$ & $+0.049$ & $+0.018$ \\ StepGame & \quad Plausibility \(P\) & $+0.418$ & $+0.197$ & $+0.373$ \\ StepGame & \quad \quad Paraphrase stability \(PS\) & $+0.186$ & $+0.249$ & $+0.112$ \\ StepGame & \quad \quad Flip consistency \(FC\) & $+0.408$ & $+0.121$ & $+0.419$ \\ \cmidrule(lr){2-5} StepGame & \textbf{Complexity} & $-0.386$ & $-0.311$ & $-0.388$ \\ StepGame & \quad Support burden \(SB\) & $-0.391$ & $-0.347$ & $-0.601$ \\ StepGame & \quad Chain length \(CL\) & $-0.371$ & $-0.214$ & $-0.395$ \\ StepGame & \quad Hard language \(HL\) & $-0.191$ & $-0.308$ & $-0.295$ \\ StepGame & \quad Diagonal burden \(DB\) & $-0.153$ & $-0.061$ & $-0.011$ \\ StepGame & \quad Selection difficulty \(SD\) & $-0.025$ & $-0.101$ & $-0.119$ \\ StepGame & \quad Entity load \(EL\) & $-0.370$ & $-0.212$ & $-0.399$ \\ \midrule SpaRTUN & \textit{Text acc.} & 67.5 & 56.4 & 72.9 \\ \cmidrule(lr){2-5} SpaRTUN & \textbf{Trustworthiness} & $+0.049$ & $+0.248$ & $+0.176$ \\ SpaRTUN & \quad Faithfulness \(F\) & $+0.030$ & $+0.182$ & $+0.132$ \\ SpaRTUN & \quad \quad Sufficiency \(S\) & $+0.089$ & $+0.276$ & $+0.182$ \\ SpaRTUN & \quad \quad Necessity \(N\) & $-0.073$ & $-0.106$ & $-0.095$ \\ SpaRTUN & \quad Plausibility \(P\) & $+0.055$ & $+0.198$ & $+0.123$ \\ SpaRTUN & \quad \quad Paraphrase stability \(PS\) & $+0.150$ & $+0.270$ & $+0.133$ \\ SpaRTUN & \quad \quad Flip consistency \(FC\) & $+0.014$ & $+0.141$ & $+0.080$ \\ \cmidrule(lr){2-5} SpaRTUN & \textbf{Complexity} & $-0.159$ & $-0.182$ & $-0.207$ \\ SpaRTUN & \quad Support burden \(SB\) & $+0.025$ & $-0.028$ & $-0.095$ \\ SpaRTUN & \quad Chain length \(CL\) & $-0.070$ & $-0.094$ & $-0.209$ \\ SpaRTUN & \quad Hard language \(HL\) & $-0.221$ & $-0.227$ & $-0.074$ \\ SpaRTUN & \quad Entity load \(EL\) & $-0.016$ & $+0.024$ & $+0.027$ \\ SpaRTUN & \quad Coreference difficulty \(CF\) & $-0.325$ & $-0.328$ & $-0.008$ \\ \bottomrule \end{tabular} \caption{Correlation between switching factors and text-only correctness. Text acc. reports exact-match accuracy. Trustworthiness factors include faithfulness \(F\), sufficiency \(S\), necessity \(N\), plausibility \(P\), paraphrase stability \(PS\), and flip consistency \(FC\). Complexity factors include support burden \(SB\), chain length \(CL\), hard language \(HL\), diagonal burden \(DB\), selection difficulty \(SD\), entity load \(EL\), and coreference difficulty \(CF\). Positive correlations indicate factors associated with higher text correctness, while negative correlations indicate factors associated with text-only failure.} \label{tab:factor_corr_alignment} \end{table*}

\subsection{Short-Circuit Computation for Switching}
\label{sec:short_circuit}

To reduce the cost of adaptive switching, we avoid computing signals that
cannot change the routing decision. Since trustworthiness is defined as
$T = 0.6F + 0.4P$ with $P \in [0,1]$, once faithfulness $F$ is known,
the final trustworthiness is bounded as
$0.6F \leq T \leq 0.6F + 0.4$.
We first compute faithfulness $F$. If the maximum possible trustworthiness
$0.6F+0.4$ is below the trust threshold $\tau_t$, the instance cannot satisfy
the trust criterion and is switched to grid-based reasoning without computing
$C$ or $P$.
Otherwise, we compute complexity $C$. If $C \geq \tau_c$, the instance is
switched to grid-based reasoning without computing $P$. If $C < \tau_c$ and
the minimum possible trustworthiness $0.6F$ already satisfies
$0.6F \geq \tau_t$, the instance remains in natural-language reasoning
without computing $P$.
Plausibility $P$ is therefore computed only when the trust bounds do not
already determine the decision. In this remaining case, we compute
$T = 0.6F + 0.4P$ and retain natural-language reasoning iff
$T \geq \tau_t$ and $C < \tau_c$; otherwise, we switch to grid-based
reasoning.
\subsection{Switching Signal Alignment}
\label{sec:switching_signal}

\subsubsection{Alignment Between Trustworthiness, Complexity, and Text Accuracy}
\label{sec:trust_complexity_alignment}

Table~\ref{tab:factor_corr_alignment} reports the correlation between each switching factor and text-only correctness on StepGame and SpaRTUN. This analysis tests whether the proposed signals identify when natural language-based reasoning is likely to succeed or fail. Since the switching policy first decides whether the text answer can be trusted, trustworthiness factors should correlate positively with text correctness, while complexity factors should correlate negatively.

On StepGame, the alignment is strong and consistent across models. Trustworthiness has a clear positive correlation with text correctness, especially through faithfulness \(F\) and its sufficiency component. This indicates that when the model's answer can be recovered from its cited support, the text answer is much more likely to be correct. Plausibility \(P\), including paraphrase stability and flip consistency, is also positive, showing that stable and logically consistent answers tend to be more reliable. Complexity shows the opposite trend: support burden, chain length, hard language, selection difficulty, and entity load are generally negatively correlated with text correctness. This is expected because StepGame is designed around controlled multi-hop spatial reasoning, where longer chains, distractor sentences, and heavier support requirements make text-only reasoning harder. Diagonal burden is weaker for some models, but still generally negative.

SpaRTUN shows a more linguistically driven pattern. Trustworthiness remains positive, especially for \textsc{LLaMA3.1-70B} and \textsc{GPT-5.1}, where faithfulness and sufficiency are strongly aligned with text correctness. The necessity component is negative across models, which is desirable: when the model can still answer after the relevant support is removed, the original text answer is less likely to be reliable. Complexity is also negative overall, but the strongest factors differ from StepGame. Chain length is less dominant, while hard language and coreference difficulty are stronger predictors of text failure. This reflects SpaRTUN's richer linguistic structure, where errors often come from resolving entity references, quantifiers, and mixed topological/directional descriptions rather than only from hop depth.

Overall, the two datasets validate different parts of the switching signal. StepGame supports the structural complexity factors, because difficulty is mainly controlled by multi-hop depth, support burden, and distractor burden. SpaRTUN supports the linguistic and grounding factors, because correctness depends more on evidence grounding, hard language, and coreference resolution. These trends justify using both trustworthiness and complexity: trustworthiness captures whether the text answer is faithful and stable, while complexity captures whether the instance itself is likely to make text-only reasoning unreliable.

{\subsubsection{Collinearity Analysis}
\label{app:factor_correlations}

We examine whether the individual trustworthiness and complexity signals
capture complementary information or exhibit substantial collinearity.
We compute pairwise Pearson correlations on the validation data. Since the
correlation matrix is symmetric, we report only its upper triangle.
The final \textbf{Err.} column reports the Pearson correlation between each factor and a binary indicator of whether the answer generated through text-only reasoning is incorrect; thus, a positive value indicates that larger values of the factor are associated with more text-only failures, whereas a negative value indicates association with fewer failures.

We exclude the aggregate Trustworthiness, Faithfulness, Plausibility, and
Complexity scores from this analysis because they are constructed from the
constituent signals and would therefore introduce expected mechanical
correlations. We also omit Entity Load (EL) for StepGame because it is
determined by hop length and is therefore redundant with Chain Length (CL)
by dataset construction.
The remaining trustworthiness components are S (Sufficiency),
N (Necessity), PS (Paraphrase Stability), and
FC (Flip Consistency). Complexity components are
SB (Support Burden), SD (Selection Difficulty),
CL (Chain Length), HL (Hard Language),
DB (Diagonal Burden), EL (Entity Load), and
CF (Coreference Difficulty).
Tables~\ref{tab:corr_large_stepgame} and \ref{tab:corr_large_spartun} report factor collinearity and error correlations for the larger models on StepGame and SpaRTUN, with each entry shown as Qwen3-32B / LLaMA3.1-70B. Tables~\ref{tab:corr_small_stepgame} and \ref{tab:corr_small_spartun} report the corresponding results for the smaller models, shown as Qwen3-8B / LLaMA3.1-8B.

% ============================================================
% LARGE -- STEPGAME
% ============================================================

\begin{table*}
\centering
\small
\renewcommand{\arraystretch}{1.05}
\resizebox{\textwidth}{!}{%
\begin{tabular}{lrrrrrrrrrr}
\toprule
 & S & N & PS & FC & SB & SD & CL & HL & DB & \textbf{Err.} \\
\midrule
S
& 1.00
& .01/.06
& .29/.41
& .62/.53
& -.63/-.47
& .10/-.04
& -.36/-.33
& -.38/-.26
& -.06/-.10
& \textbf{-.48/-.37} \\

N
&
& 1.00
& -.03/.02
& .11/-.02
& .03/.07
& -.05/-.11
& -.04/-.12
& -.05/.04
& .09/-.05
& .06/-.12 \\

PS
&
&
& 1.00
& .78/.55
& -.67/-.52
& .13/.01
& -.08/-.25
& -.31/-.22
& -.18/.02
& \textbf{-.28/-.41} \\

FC
&
&
&
& 1.00
& -.57/-.46
& .05/.17
& -.11/-.10
& -.26/-.22
& -.13/.09
& -.18/-.32 \\

SB
&
&
&
&
& 1.00
& -.24/-.20
& .44/.48
& .48/.45
& .15/.04
& \textbf{.43/.41} \\

SD
&
&
&
&
&
& 1.00
& .70/.72
& -.02/.07
& .09/.12
& -.09/.15 \\

CL
&
&
&
&
&
&
& 1.00
& .33/.35
& .15/.12
& .22/.39 \\

HL
&
&
&
&
&
&
&
& 1.00
& .25/.15
& .18/.35 \\

DB
&
&
&
&
&
&
&
&
& 1.00
& .11/.09 \\
\bottomrule
\end{tabular}}
\caption{Pairwise Pearson correlations on StepGame for
Qwen3-32B / LLaMA3.1-70B.
S: Sufficiency; N: Necessity; PS: Paraphrase Stability;
FC: Flip Consistency; SB: Support Burden;
SD: Selection Difficulty; CL: Chain Length;
HL: Hard Language; DB: Diagonal Burden.
Err. denotes correlation with the binary text-only error indicator.}
\label{tab:corr_large_stepgame}
\end{table*}

% ============================================================
% LARGE -- SPARTUN
% ============================================================

\begin{table*}
\centering
\small
\renewcommand{\arraystretch}{1.05}
\resizebox{\textwidth}{!}{%
\begin{tabular}{lrrrrrrrrr}
\toprule
 & S & N & PS & FC & CL & HL & EL & CF & \textbf{Err.} \\
\midrule
S
& 1.00
& -.03/-.04
& .17/.31
& .18/.19
& -.15/-.12
& -.15/-.22
& -.04/.06
& -.12/-.23
& \textbf{-.27/-.35} \\

N
&
& 1.00
& -.10/-.07
& .02/.02
& .12/.12
& .08/.07
& .05/.03
& .11/.09
& .08/.08 \\

PS
&
&
& 1.00
& .14/.23
& -.13/-.08
& -.09/-.22
& .01/.07
& -.14/-.23
& \textbf{-.29/-.34} \\

FC
&
&
&
& 1.00
& -.16/-.03
& -.08/-.21
& -.07/.03
& -.09/-.22
& -.15/-.23 \\

CL
&
&
&
&
& 1.00
& .11/.14
& .09/.13
& .08/.17
& .16/.06 \\

HL
&
&
&
&
&
& 1.00
& .45/-.07
& .54/.53
& .16/.41 \\

EL
&
&
&
&
&
&
& 1.00
& .39/.09
& .12/-.13 \\

CF
&
&
&
&
&
&
&
& 1.00
& \textbf{.27/.48} \\
\bottomrule
\end{tabular}}
\caption{Pairwise Pearson correlations on SpaRTUN for
Qwen3-32B / LLaMA3.1-70B.
S: Sufficiency; N: Necessity; PS: Paraphrase Stability;
FC: Flip Consistency; CL: Chain Length; HL: Hard Language;
EL: Entity Load; CF: Coreference Difficulty.
Err. denotes correlation with the binary text-only error indicator.}
\label{tab:corr_large_spartun}
\end{table*}
% ============================================================
% SMALL -- STEPGAME
% ============================================================

\begin{table*}
\centering
\small
\renewcommand{\arraystretch}{1.05}
\resizebox{\textwidth}{!}{%
\begin{tabular}{lrrrrrrrrrr}
\toprule
 & S & N & PS & FC & SB & SD & CL & HL & DB & \textbf{Err.} \\
\midrule
S
& 1.00
& -.01/.01
& .19/.27
& .57/.22
& -.63/-.14
& -.03/-.17
& -.29/-.21
& -.30/-.17
& -.02/-.19
& \textbf{-.58/-.14} \\

N
&
& 1.00
& -.04/.09
& -.05/-.06
& .03/.07
& -.08/.04
& -.04/.08
& .09/.05
& .01/-.04
& -.08/.13 \\

PS
&
&
& 1.00
& .38/.07
& -.22/-.48
& .12/.02
& .05/-.23
& -.17/-.25
& -.01/.01
& -.12/-.28 \\

FC
&
&
&
& 1.00
& -.52/-.20
& .31/.57
& .18/.41
& -.18/-.22
& -.02/-.43
& -.39/-.19 \\

SB
&
&
&
&
& 1.00
& -.08/-.03
& .35/.48
& .33/.41
& .01/.04
& \textbf{.52/.08} \\

SD
&
&
&
&
&
& 1.00
& .65/.63
& .08/.07
& .06/-.04
& .15/.21 \\

CL
&
&
&
&
&
&
& 1.00
& .20/.28
& .05/-.01
& .31/.21 \\

HL
&
&
&
&
&
&
&
& 1.00
& .22/.21
& .23/.09 \\

DB
&
&
&
&
&
&
&
&
& 1.00
& .08/.09 \\
\bottomrule
\end{tabular}}
\caption{Pairwise Pearson correlations on StepGame for
Qwen3-8B / LLaMA3.1-8B.
S: Sufficiency; N: Necessity; PS: Paraphrase Stability;
FC: Flip Consistency; SB: Support Burden;
SD: Selection Difficulty; CL: Chain Length;
HL: Hard Language; DB: Diagonal Burden.
Err. denotes correlation with the binary text-only error indicator.}
\label{tab:corr_small_stepgame}
\end{table*}

% ============================================================
% SMALL -- SPARTUN
% ============================================================
\begin{table*}
\centering
\small
\renewcommand{\arraystretch}{1.05}
\resizebox{\textwidth}{!}{%
\begin{tabular}{lrrrrrrrrr}
\toprule
 & S & N & PS & FC & CL & HL & EL & CF & \textbf{Err.} \\
\midrule
S
& 1.00
& .02/-.02
& .27/.27
& .06/.10
& -.10/-.04
& -.01/-.19
& .06/.01
& -.05/-.32
& \textbf{-.13/-.26} \\

N
&
& 1.00
& -.03/-.04
& -.01/.01
& .03/.01
& -.00/-.05
& -.03/.03
& -.01/-.03
& .05/-.09 \\

PS
&
&
& 1.00
& .02/.20
& -.10/-.13
& .01/-.29
& .11/-.04
& -.02/-.41
& \textbf{-.25/-.34} \\

FC
&
&
&
& 1.00
& -.09/-.11
& -.00/-.22
& .03/-.02
& -.02/-.31
& -.16/-.26 \\

CL
&
&
&
&
& 1.00
& .06/.11
& .02/.22
& .05/.09
& .16/.08 \\

HL
&
&
&
&
&
& 1.00
& .58/.12
& .61/.57
& .15/.37 \\

EL
&
&
&
&
&
&
& 1.00
& .59/.09
& .08/.22 \\

CF
&
&
&
&
&
&
&
& 1.00
& \textbf{.12/.44} \\
\bottomrule
\end{tabular}}
\caption{Pairwise Pearson correlations on SpaRTUN for
Qwen3-8B / LLaMA3.1-8B.
S: Sufficiency; N: Necessity; PS: Paraphrase Stability;
FC: Flip Consistency; CL: Chain Length; HL: Hard Language;
EL: Entity Load; CF: Coreference Difficulty.
Err. denotes correlation with the binary text-only error indicator.}
\label{tab:corr_small_spartun}
\end{table*}

\paragraph{Observed trends.}
Several patterns emerge from the correlation analysis. First,
trustworthiness-related and complexity-related components often show
opposite associations. On StepGame, for example, Sufficiency and
Paraphrase Stability are negatively correlated with several complexity
factors, particularly Support Burden. This is consistent with the intuition
behind our routing rule, which favors text-only reasoning when
trustworthiness is high and complexity is low. Necessity behaves
differently, showing mostly weak or mixed correlations with the other
components, suggesting that it serves as a complementary
signal to sufficiency.

The error correlations further show that the different factors are
informative in different settings rather than behaving as a single
universal difficulty signal. Trustworthiness components such as
Sufficiency and Paraphrase Stability are generally associated with fewer
text-only errors, while several complexity components are associated with
more errors. However, the strength of these correlations change across
datasets and model sizes. For example, on SpaRTUN, Hard Language and
Coreference Difficulty are more strongly associated with error for
LLaMA3.1-8B than for Qwen3-8B, whereas on StepGame Support Burden is
substantially more informative for Qwen3-8B. Similar shifts are also
visible among the larger models. This suggests that different models
struggle with different aspects of spatial reasoning, and that the most
useful signal depends on both the model and the dataset. Using multiple
complementary factors therefore makes the switching criterion more
adaptable to changes in model capability and problem setting, rather than
relying on a single fixed notion of difficulty.

Finally, most pairwise correlations are low to moderate, although some
stronger dataset-specific relationships occur. Selection Difficulty and
Chain Length are strongly correlated on StepGame, while Hard Language and
Coreference Difficulty show stronger overlap on SpaRTUN. Importantly,
these relationships are not consistent across all models and datasets,
which further indicates that the individual factors are not interchangeable.
We therefore retain multiple trustworthiness and complexity signals to
capture complementary sources of reasoning difficulty across datasets and
model scales.

\subsubsection{Human Evaluation of Switching Signals}
\label{app:human_evaluation}

\paragraph{Data and assignment.}
We sample $n{=}240$ instances from the switching experiments, with $n{=}60$ instances for
each model and dataset combination. Three annotators judge the samples that cover different reasoning
depths and include the story, question, model answer, and support sentences
selected by the model.

\paragraph{Procedure.}
The evaluation has two phases. First, annotators view only the story and
question and rate complexity. The model output remains hidden so that it
cannot influence the perceived difficulty. Annotators then reveal the model
answer and support sentences and rate faithfulness, plausibility, and
trustworthiness. They are instructed to use only the provided story,
question, answer, and support sentences. They should not use outside knowledge
or judge an answer based on its fluency, the model identity, or how confident
it sounds.

\paragraph{Complexity.}
Annotators consider the number of entities and relations, reasoning depth,
distractors, coreference, nested relations, and difficult language.

\begin{enumerate}
    \item A single direct relation with few entities and no distractors,
    coreference, or nested relations.
    \item One or two simple steps with clear references and little distracting
    or difficult language.
    \item Multiple relations with some distractors, nested relations, coreference, or
    careful wording.
    \item Several relations and entities with substantial confusion risk from
    text alone.
    \item A long chain with many entities, nested or coreferential references,
    several distractors, or difficult language. A grid or diagram would
    likely be needed.
\end{enumerate}

\paragraph{Faithfulness.}
Annotators assess whether the selected support sentences are relevant,
complete, and sufficient for deriving the answer.

\begin{enumerate}
    \item The support is irrelevant or contradictory, or the answer cannot be
    derived.
    \item Some relevant evidence is present, but major reasoning links are
    missing.
    \item The support captures part of the answer but not the complete stated
    relation.
    \item Nearly all required evidence is present, with a small missing link
    or minor incompleteness.
    \item All required evidence is present, relevant, and correctly connected.
\end{enumerate}

\paragraph{Plausibility.}
Annotators assess whether the reasoning is logically consistent with the
story, including relation direction and entity order.

\begin{enumerate}
    \item The reasoning directly contradicts the story or contains a severe
    logical error.
    \item The reasoning contains a major error in composing different relations together, with only a small part showing clear understanding of the story.
    \item Some relations are correct, but the final answer is incomplete or
    reflects partial confusion.
    \item The main spatial relation is correct, with only a minor
    inconsistency or ambiguity.
    \item All relevant relations are correctly composed, entity order is
    preserved, and nothing unsupported is added.
\end{enumerate}

\paragraph{Trustworthiness.}
Annotators assess whether the response can be trusted based on its evidence,
logic, and completeness.

\begin{enumerate}
    \item The response should definitely not be trusted because it contains a
    clear error, contradiction, unsupported conclusion, or major missing
    step.
    \item The response has substantial problems and should be checked before
    use.
    \item The response has mixed strengths and weaknesses or insufficient
    evidence to decide.
    \item The response is well supported and coherent, with only minor
    residual uncertainty.
    \item The response is fully supported, consistent, and complete, with no
    further checking required.
\end{enumerate}

\paragraph{Experiment Design.}
Human ratings range from 1 to 5, while the switching signals are continuous
values in $[0,1]$. Also, the complexity signal uses the saturating function for frequency-based factors, which compresses high values while preserving
their order. We therefore report Spearman's $\rho$, which measures 
rank alignment without requiring comparable score magnitudes using mean of the three human ratings. We compute ordinal
Krippendorff's $\alpha$ for inter-annotator agreement scores.

\section{Modalities in Small Language Models}
\label{app:small_lm_modalities}

\textbf{Behavior of Small Language Models.}

Table~\ref{tab:accuracy_khop_stepgame_appendix} compares smaller open-source models on StepGame across text, relational triples, coordinates, grids, and ToT-CoT prompting. The results demonstrate that structured spatial representations become more useful as hop length increases. Text-only reasoning is competitive for short chains, especially at one hop, but drops as the model must compose multiple relations for reasoning. Relational triples make the extracted relations explicit but still require internal symbolic composition. In contrast, grids externalize the spatial layout, making them the most stable representation overall. Pruned grids further reduce irrelevant entities and keep the reasoning context compact, while full grids preserve the full scene but may force smaller models to reason over unnecessary entities. This matters less for stronger reasoning models: Qwen models show similar full-grid and pruned-grid performance, whereas LLaMA3.1-8B benefits more clearly from pruning. Coordinate representations also help, but their reliance on positional comparison and numerical or symbolic manipulation makes them less consistently reliable than grids.

Table~\ref{tab:spartun_resq_results1} extends the analysis to SpaRTUN, which includes both Yes/No and Find-Relation questions. Here, grid-based methods are not always dominant because they depend on the quality of upstream relation extraction. For LLaMA3.1-8B, CoS performs best, suggesting that very weak models may benefit more from guided symbolic reasoning than from grids built from noisy extracted relations. For Qwen3-8B, performance varies by question type: CoS is strongest on YN questions, relational triples perform best on FR and overall accuracy, while  pruned grids remain competitive. For Qwen3-14B, pruned grids perform best overall and are especially strong on Find-Relation questions, improving overall accuracy from 64.90\% to 72.97\%. This suggests that compact structured layouts become more useful when the model can extract relations more reliably.

To test generalization beyond StepGame and SpaRTUN, we also evaluate ReSQ, where grids are constructed using GPT-5.1 from relations extracted by each model. Since ReSQ contains shorter Yes/No spatial reasoning chains, gains from grid grounding are smaller than on StepGame but still model-dependent. For LLaMA3.1-8B, text-only reasoning remains strongest. In contrast, Qwen3-8B and Qwen3-14B benefit from grid grounding, with full or pruned grids giving the best accuracy on ReSQ. This suggests that when relation extraction is reliable enough, grid-based layouts help focus the model on the spatial evidence relevant to the question. Relational triples show less consistent gains, making them useful in some cases but less reliable than grids for ReSQ.

\begin{table*}[h!]

    \centering

    \small

    \renewcommand{\arraystretch}{1.2}

    \setlength{\tabcolsep}{5pt}

    \begin{tabular}{lccccccccccc}

    \toprule

    \textbf{Model / Setting} & \textbf{1} & \textbf{2} & \textbf{3} & \textbf{4} & \textbf{5} & \textbf{6} & \textbf{7} & \textbf{8} & \textbf{9} & \textbf{10} & \textbf{Overall} \\

    \midrule

    \textbf{LLaMA3.1-8B Text}              & 42 & 25 & 22 & 19 & 18 & 16 & 15 & 14 & 13 & 13 & 19.8 \\

    \textbf{LLaMA3.1-8B Relational Triples}  & \textbf{53} & 29 & 22 & 22 & 17 & 15 & 17 & 14 & 12 & 12 & 21.3 \\

    \textbf{LLaMA3.1-8B Pruned Coordinate} & 34 & \underline{32} & 34 & 25 & \underline{25} & \underline{24} & \underline{25} & \underline{24} & \underline{23} & 21 & \underline{26.6} \\

    \textbf{LLaMA3.1-8B Full Coordinate}   & 32 & 29 & 28 & \underline{26} & \underline{25} & 23 & 22 & 20 & 20 & 17 & 24.3 \\

    \textbf{LLaMA3.1-8B Pruned Grid}       & 40 & \textbf{35} & \underline{35} & \textbf{34} & \textbf{31} & \textbf{28} & \textbf{32} & \textbf{35} & \textbf{34} & \textbf{30} & \textbf{33.3} \\

    \textbf{LLaMA3.1-8B Full Grid}         & 36 & 25 & 28 & \underline{26} & 20 & 22 & 21 & \underline{24} & \underline{23} & \underline{23} & 24.8 \\

    \textbf{LLaMA3.1-8B ToT-CoT}           & \underline{45} & 31 & \textbf{36} & 24 & \underline{25} & \underline{24} & 20 & 20 & 21 & 20 & \underline{26.6} \\

    \midrule

    \textbf{Qwen3-8B Text}                 & \underline{95} & 69 & 70 & 64 & 58 & 57 & 53 & 42 & 42 & 41 & 59.2 \\

    \textbf{Qwen3-8B Relational Triples}   & 94 & 80 & 69 & 60 & 59 & 58 & 62 & 42 & 43 & 38 & 60.5 \\

    \textbf{Qwen3-8B Pruned Coordinate}    & \textbf{96} & \underline{88} & \textbf{93} & \underline{78} & 62 & 59 & \underline{64} & \underline{64} & 60 & \underline{62} & 72.6 \\

    \textbf{Qwen3-8B Full Coordinate}      & 94 & 84 & \underline{89} & 76 & 63 & \underline{63} & 63 & 61 & 60 & 59 & 71.2 \\

    \textbf{Qwen3-8B Pruned Grid}          & 88 & \textbf{91} & 84 & \textbf{84} & \textbf{85} & \textbf{85} & \textbf{87} & \textbf{83} & \underline{80} & \textbf{80} & \textbf{84.8} \\

    \textbf{Qwen3-8B Full Grid}            & 87 & \underline{88} & 83 & \textbf{84} & \underline{84} & \textbf{85} & \textbf{87} & \textbf{83} & \textbf{82} & \textbf{80} & \underline{84.4} \\

    \textbf{Qwen3-8B ToT-CoT}              & 79 & 58 & 61 & 61 & 51 & 48 & 46 & 35 & 37 & 36 & 51.3 \\

    \midrule

    \textbf{Qwen3-14B Text}                & \textbf{98} & 81 & 73 & 71 & 64 & 62 & 63 & 58 & 56 & 52 & 67.8 \\

    \textbf{Qwen3-14B Relational Triples}  & \underline{89} & 79 & 73 & 68 & 61 & 56 & 58 & 51 & 48 & 47 & 63.0 \\

    \textbf{Qwen3-14B Pruned Coordinate}   & 85 & 83 & 82 & \underline{81} & \underline{77} & \underline{76} & \underline{81} & 76 & \underline{78} & 76 & 79.5 \\

    \textbf{Qwen3-14B Full Coordinate}     & 83 & 78 & 78 & 74 & 73 & 71 & 70 & 72 & 69 & 71 & 73.9 \\

    \textbf{Qwen3-14B Pruned Grid}         & \underline{89} & \textbf{89} & \textbf{87} & \textbf{84} & \textbf{84} & \textbf{82} & \textbf{85} & \underline{80} & \textbf{82} & \underline{79} & \underline{84.2} \\

    \textbf{Qwen3-14B Full Grid}           & \underline{89} & \underline{87} & \underline{86} & \textbf{84} & \textbf{84} & \textbf{82} & \textbf{85} & \textbf{84} & \textbf{82} & \textbf{80} & \textbf{84.3} \\

    \textbf{Qwen3-14B ToT-CoT}             & \underline{89} & 76 & 70 & 62 & 57 & 54 & 57 & 55 & 52 & 52 & 62.5 \\

    \bottomrule

    \end{tabular}

    \caption{Accuracy (\%) across k-hop levels on StepGame for text-, relation-triple-, grid-, coordinate-, and ToT-CoT-based representations. Each k-hop column is computed over $n{=}500$ instances, and Overall is computed over $n{=}5{,}000$ instances. Best results are bolded and second-best results are underlined within each model block.}

    \label{tab:accuracy_khop_stepgame_appendix}

\end{table*}

\begin{table}[t]

\vspace{-6pt}

\centering

\tiny

\setlength{\tabcolsep}{2.0pt}

\renewcommand{\arraystretch}{0.94}

\resizebox{\columnwidth}{!}{%

\begin{tabular}{lcccc}

\toprule

& \multicolumn{3}{c}{\textbf{SpaRTUN}} & \textbf{ReSQ} \\

\cmidrule(lr){2-4}\cmidrule(lr){5-5}

\textbf{Model / Setting} & \textbf{YN} & \textbf{FR} & \textbf{Overall} & \textbf{Overall} \\

\midrule

LLaMA3.1-8B Text       

& 58.50 & 11.80 & \underline{37.25} & \textbf{58.60} \\

LLaMA3.1-8B Rel. Trip. 

& 58.20 & 11.40 & 36.91 & 57.50 \\

LLaMA3.1-8B CoS        

& \textbf{68.09} & \textbf{17.06} & \textbf{44.87} & -- \\

LLaMA3.1-8B FG         

& \underline{59.08} & 10.20 & 36.84 & 57.87 \\

LLaMA3.1-8B PG         

& 57.77 & \underline{12.16} & 37.02 & \underline{58.03} \\

\midrule

Qwen3-8B Text          

& 74.20 & 43.40 & 60.19 & 75.57 \\

Qwen3-8B Rel. Trip.    

& 72.18 & \textbf{51.96} & \textbf{62.98} & 70.98 \\

Qwen3-8B CoS           

& \textbf{85.76} & 30.98 & 60.84 & -- \\

Qwen3-8B FG            

& 71.69 & 41.57 & 57.98 & \underline{79.84} \\ 

Qwen3-8B PG            

& \underline{74.96} & \underline{44.12} & \underline{60.93} & \textbf{80.33} \\

\midrule

Qwen3-14B Text         

& 82.00 & 44.68 & 64.90 & 77.21 \\

Qwen3-14B Rel. Trip.   

& 73.16 & 56.67 & 65.60 & 73.61 \\

Qwen3-14B CoS          

& \textbf{83.80} & 39.80 & 63.78 & -- \\

Qwen3-14B FG           

& 77.09 & \underline{59.02} & \underline{68.87} & \textbf{83.28} \\

Qwen3-14B PG           

& \underline{82.32} & \textbf{61.76} & \textbf{72.97} & \underline{82.13} \\

\bottomrule

\end{tabular}%

}

\caption{Accuracy (\%) on SpaRTUN and ReSQ. SpaRTUN reports Yes/No (YN), Find Relations (FR), and overall accuracy. ReSQ reports overall Yes/No accuracy. For ReSQ, grids are constructed using GPT-5.1 and then used by the QA model. FG and PG denote full-grid and pruned-grid reasoning, respectively; Rel. Trip. denotes relational-triple reasoning. Best and second-best results are bolded and underlined within each model block.}

\label{tab:spartun_resq_results1}

\vspace{-8pt}

\end{table}

\section{Switching on Small Language Models}

\begin{table}[t]
% \vspace{-6pt}
\centering
\small
\setlength{\tabcolsep}{3.2pt}
\renewcommand{\arraystretch}{0.95}
\resizebox{\columnwidth}{!}{%
\begin{tabular}{lcccc|cccc}
\toprule
\multirow{2}{*}{\textbf{Setting}} 
& \multicolumn{4}{c|}{\textbf{Acc.}} 
& \multicolumn{4}{c}{\textbf{Rate}} \\
\cmidrule(lr){2-5}\cmidrule(lr){6-9}
& \textbf{Txt} & \textbf{G} & \textbf{AS} & \textbf{Orc}
& \textbf{S} & \textbf{T} & \textbf{C} & \textbf{B} \\
\midrule
 Qwen3-8B    
& 52.2 & 87.5 & 87.2 & 91.7 & 80.3 & 12.0 & 24.0 & 64.0 \\
 Qwen3-14B   
& 63.3 & 87.1 & 88.0 & 93.3 & 65.6 & 58.0 & 9.0  & 33.0 \\
 LLaMA3.1-8B 
& 20.2 & 28.1 & 27.1 & 32.5 & 94.0 & 22.0 & 13.0 & 64.0 \\
\bottomrule
\end{tabular}%
}
\caption{Accuracy and switching behavior on StepGame. Results are reported on the StepGame test set with $n{=}1{,}000$ instances per hop. Txt denotes text-only accuracy, G denotes grid-only accuracy, AS denotes adaptive switching accuracy, and Orc denotes oracle accuracy. S is the overall switch rate; T, C, and B denote the percentage of switched instances attributed to low trustworthiness, high complexity, and both signals, respectively.}
\label{tab:overall_accuracy_stepgame_small}
\vspace{-8pt}
\end{table}

Table~\ref{tab:overall_accuracy_stepgame_small} shows adaptive switching remains close to or improves over grid-only performance for small open-source models. This indicates that the switching policy can preserve the benefit of structured reasoning while still selecting text when it is reliable enough. 
The switch rate also reflects model strength. Weaker models such as LLaMA3.1-8B switch most often, with a 94.0\% switch rate, suggesting that their text-only answers are rarely reliable enough for multi-hop spatial reasoning. In contrast, Qwen3-14B switches less often at 65.6\% and achieves the best adaptive accuracy. This suggests that stronger reasoning models can rely on text-based reasoning more selectively while still benefiting from grid-based reasoning on harder or less trustworthy instances. The remaining gap to oracle accuracy shows that better modality selection could further improve performance, especially in deciding when text is sufficient and when grid grounding is necessary.

\section{Ablation Study}
\label{app:ablation}

We conduct ablation studies to isolate the contribution and reliability of each component in our structured reasoning pipeline. First, we evaluate ground-truth intermediate representations to estimate the upper bound of relational triples, full grids, and pruned grids when upstream extraction noise is removed. Second, we analyze construction reliability by separating errors from relation extraction and errors introduced during grid construction. Third, we evaluate the spatial relation extraction module itself, comparing the baseline extraction setting with our sentence-level, context-aware extraction pipeline. Finally, we analyze adaptive routing cost and signal-specific contribution by comparing trust-only, complexity-only, and full switching policies. Together, these ablations show where performance gains come from, when grid-based reasoning is limited by upstream extraction, and how switching balances accuracy with computational cost.

\subsection{Ground-Truth Representation Upper Bound} 
To isolate upstream relation extraction errors, we evaluate reasoning directly on ground-truth intermediate representations in Table \ref{tab:gt_upper_bound_appendix}. Specifically, we compare reasoning over ground-truth relational triples, ground-truth full grids, and ground-truth pruned grids. Ground-truth full grids are constructed directly from ground-truth relational triples using our grid-construction pipeline, while pruning is performed model-wise before applying the same construction code. We use model-wise pruning rather than dataset-provided reasoning annotations because the annotations are Prolog-based and answer-oriented, and do not translate directly into the entity set needed for grid construction. This is especially important for Find-Relations multi-label questions, where pruning must retain the queried entities and enough surrounding context to preserve the answer. Model-wise pruning is therefore more consistent with the deployed pipeline, remains dataset-agnostic, and attributes pruning errors to the same model used for reasoning. Ground-truth triples are passed directly to the model without any spatial layout construction. This analysis measures the upper bound of each representation when extraction noise is removed.

\begin{table}[t]
\centering
\small
\setlength{\tabcolsep}{3pt}
\renewcommand{\arraystretch}{0.95}
\begin{tabular}{llcc}
\toprule
\textbf{Model} & \textbf{GT Rep.} & \textbf{SpaRTUN} & \textbf{StepGame} \\
\midrule
LLaMA3.1-70B & Rel. Trip. & 62.92 & 45.00 \\
LLaMA3.1-70B & FG         & 65.24 & 88.90 \\
LLaMA3.1-70B & PG         & \textbf{65.42} & \textbf{97.20} \\
\midrule
Qwen3-32B & Rel. Trip. & 84.14 & 72.00 \\
Qwen3-32B & FG         & \textbf{89.13} & \textbf{100.00} \\
Qwen3-32B & PG         & \underline{88.77} & \underline{99.50} \\
\bottomrule
\end{tabular}
\caption{Ground-truth upper bound analysis. Rel. Trip., FG, and PG denote ground-truth relational triples, full grids, and pruned grids, respectively.}
\label{tab:gt_upper_bound_appendix}
\end{table}

Ground-truth grids substantially outperform relational triples, especially on both datasets where deeper multi-hop spatial reasoning benefits from explicit structured layouts. This suggests that even perfect qualitative relational triples are often insufficient on their own and that models benefit from representations that combine qualitative relations with an explicit quantitative spatial structure. 

For Qwen3-32B, pruning remains nearly identical to full grids on both SpaRTUN and StepGame, likely because the model already performs strongly on structured reasoning tasks, consistent with observations in the Qwen3 technical report~\citep{yang2025qwen3technicalreport}. On SpaRTUN, pruning itself is also a harder task because many Yes/No questions contain quantifiers while Find Relation questions often require complete block-level reasoning. As a result, the pruned and full grids produce identical answers for 91.6\% of examples overall on a subset of $n{=}561$ instances (50\% of our test set while maintaining the same split proportions), with agreement reaching 95.8\% for YN questions and 86.7\% for FR questions. This suggests that much of the useful spatial structure in SpaRTUN must often be preserved even after pruning.

\subsection{Construction Reliability}
\label{sec:construction_reliability}

Since the grid-based route depends on the quality of the intermediate structure, we separately analyze the reliability of relation extraction and grid construction. This section reports how accurately spatial relations are extracted from text and how reliably those relations are converted into valid grid representations for downstream reasoning.
\begin{table*}[t]
\centering
\small
\setlength{\tabcolsep}{3.2pt}
\renewcommand{\arraystretch}{0.95}
\begin{tabular}{llccc|ccc|ccc}
\toprule
\multirow{2}{*}{\textbf{Dataset}} 
& \multirow{2}{*}{\textbf{Model}} 
& \multicolumn{3}{c|}{\textbf{Relation Extraction}} 
& \multicolumn{3}{c|}{\textbf{Gold Rel. with Grid}} 
& \multicolumn{3}{c}{\textbf{Pred. Rel. with Grid}} \\
\cmidrule(lr){3-5}\cmidrule(lr){6-8}\cmidrule(lr){9-11}
& & \textbf{P} & \textbf{R} & \textbf{F1} 
& \textbf{P} & \textbf{R} & \textbf{F1} 
& \textbf{P} & \textbf{R} & \textbf{F1} \\
\midrule
\multirow{3}{*}{SpaRTUN} 
& Qwen3-32B    
& 81.3 & 79.4 & 80.3 
& 91.24 & 99.42 & 95.16 
& 69.0 & 76.15 & 72.40 \\
& LLaMA3.1-70B 
& 79.0 & 76.1 & 77.5 
& 91.24 & 99.42 & 95.16 
& 70.7 & 75.45 & 73.00 \\
& GPT-5.1      
& 88.8 & 86.2 & 87.5
& 91.24 & 99.42 & 95.16 
& 75.84 & 81.11 & 78.39 \\
\midrule
\multirow{3}{*}{StepGame} 
& Qwen3-32B    
& 96.6 & 92.1 & 94.3 
& 100.00 & 100.00 & 100.00 
& 88.54 & 90.85 & 89.68 \\
& LLaMA3.1-70B 
& 98.5 & 91.1 & 94.7 
& 100.00 & 100.00 & 100.00 
& 85.25 & 90.63 & 87.86 \\
& GPT-5.1      
& 97.3 & 92.8 & 95.0 
& 100.00 & 100.00 & 100.00 
& 84.19 & 89.32 & 86.68 \\
\bottomrule
\end{tabular}
\caption{Reliability of relation extraction and grid construction. Relation Extraction evaluates how many predicted spatial relations match the gold relation set. Gold Rel. with Grid evaluates grid encoding fidelity when the grid is constructed from gold relations, isolating the grid construction step from upstream extraction errors. Pred. Rel. with Grid evaluates the full predicted pipeline, where extracted relations are first generated by the model and then encoded into a grid.}
\label{tab:merged-rel-grid-results}
\end{table*}

\subsubsection{Relation extraction and grid construction fidelity.} 
Table~\ref{tab:merged-rel-grid-results} evaluates two sources of error in the structured pipeline: relation extraction and grid construction. Relation extraction is evaluated as a relation-level matching problem, measuring how many predicted spatial triples match the gold triples. Grid construction is different: it evaluates whether a complete grid layout faithfully encodes the target relation graph. In this setting, recall measures how many intended relations are preserved in the grid, while precision can decrease if the concrete placement of objects induces additional spatial relations that were not explicitly specified in the ground-truth relations.

This distinction matters most for \textsc{SpaRTUN}. Unlike \textsc{StepGame}, which mainly contains controlled directional relations, \textsc{SpaRTUN} includes directional, topological, distance-based, and RCC-style relations. Encoding all of these constraints into a single discrete row-column grid is a harder geometric realization problem and can become computationally difficult in the presence of mixed spatial constraints. Even when the ground-truth relations are used, the grid achieves very high recall but slightly lower precision, because assigning each entity a concrete grid position can introduce extra directional cues between pairs that are not explicitly constrained in the original relation graph. Thus, the grid should be viewed as a compact reasoning interface rather than a perfect reconstruction of every spatial relation.

The ground-truth relations results show that the grid construction procedure itself is reliable: it is perfect on \textsc{StepGame} and preserves almost all gold relations on \textsc{SpaRTUN}. When predicted relations are used, performance decreases, especially on \textsc{SpaRTUN}, showing that most remaining errors come from upstream relation extraction and from the difficulty of realizing richer spatial graphs in a single grid. This supports our design choice: the goal is not to solve exact geometric realization for every spatial relation in a story but to test whether a structured grid representation can still provide a useful intermediate reasoning interface for LLMs.

\subsubsection{Spatial Relation Extraction}
\label{sec:relation_extraction}

Grid construction relies heavily on the quality of the extracted spatial
relations. We therefore evaluate relation extraction as a preprocessing step.
The task converts each spatial story into triples of the form
\texttt{(head, relation, tail)}, which are then passed to the
grid-construction module.

Our extraction pipeline processes the story sentence by sentence rather than
extracting all relations in a single pass. For StepGame, we use a line-by-line
extraction module with in-context learning, allowing the model to focus on
local directional relations. For SpaRTUN, we use the same sentence-level
extraction followed by a coreference-aware refinement pass, where each line is
reprocessed using the previous two lines and the already extracted relations
as context. This is particularly useful for SpaRTUN because its stories contain
richer language, pronouns, and entity references that require additional
grounding and coreference resolution.

To assess the benefit of this decomposition, we compare our pipeline against a
baseline that provides the full story in a single call and asks the model to
extract all spatial relations jointly. The baseline must simultaneously handle
relation extraction, entity grounding, and coreference, which can lead to
missed or noisy triples. In contrast, our sentence-level extraction and
context-aware refinement decompose the task into smaller decisions. As shown
in Table~\ref{tab:relation_extraction}, this improves relation-extraction F1
from 60.0 to 80.3 on SpaRTUN and from 69.0 to 94.3 on StepGame. These gains
are important because relation-extraction errors propagate directly to grid
construction and downstream reasoning.

\begin{table}[t]
\centering
\small
\setlength{\tabcolsep}{4pt}
\renewcommand{\arraystretch}{0.95}
\resizebox{\columnwidth}{!}{%
\begin{tabular}{lcc}
\toprule
\textbf{Dataset / Model} & \textbf{Baseline} & \textbf{Rel. Extr. Pipeline} \\
\midrule
SpaRTUN / Qwen3-32B  & 60.0 & 80.3 \\
StepGame / Qwen3-32B & 69.0 & 94.3 \\
\bottomrule
\end{tabular}%
}
\caption{Relation-extraction F1 before grid construction. Baseline extracts
all relations directly from the full story in a single pass, while Rel. Extr.
Pipeline denotes our sentence-level relation extraction pipeline.}
\label{tab:relation_extraction}
\end{table}

\subsection{Cost Analysis}
\label{sec:rq3_cost_analysis}

Table~\ref{tab:routing_costs} reports the average token cost per record for adaptive routing across StepGame and SpaRTUN. We compare the fixed text-only route, the fixed grid pipeline, the cost of computing each switching signal, and the final adaptive route. The switch total includes both the switching-signal cost and the cost of the selected reasoning path.

The main observation is that adaptive switching changes the cost--accuracy tradeoff differently across datasets. On StepGame, adaptive switching improves or preserves accuracy relative to the fixed grid route, while token cost varies by model: it decreases slightly for \textsc{Qwen3-32B}, remains close for \textsc{LLaMA3.1-70B}, and increases for \textsc{GPT-5.1}. This mixed pattern indicates that, for StepGame, the switching signal is useful mainly as a routing and accuracy mechanism rather than as a guaranteed token-saving mechanism. Since the grid pipeline is already highly effective on this dataset, the additional cost of computing the switch decision can offset the savings from routing some examples through the text path.

SpaRTUN shows a different pattern. Here, adaptive switching reduces average token cost relative to always using the grid pipeline across all three models, while keeping accuracy competitive with the stronger fixed route. This suggests that the routing policy is more cost-effective when the dataset contains a mixture of examples, some that benefit from structured reasoning and others that can remain on the cheaper text-only path. The reduction is largest for \textsc{GPT-5.1}; however, because \textsc{GPT-5.1} also has lower costs across most fixed and adaptive rows, we interpret this as partly a model-specific efficiency effect rather than only an effect of routing.
The single-signal rows help explain why the combined policy is preferable. In these ablations, one signal is disabled while the other remains active: trust-only routing zeroes out the complexity signal and uses the trust threshold $\tau_t$, while complexity-only routing zeroes out the trustworthiness signal and uses the complexity threshold $\tau_c$. Each signal provides useful but incomplete information. Trustworthiness captures whether the model's current text answer appears grounded and stable, while complexity captures factors that make the instance difficult before or during reasoning. As shown in Table~\ref{tab:factor_corr_alignment}, these factors correlate with text-only correctness in the expected directions: trust-related factors are mostly positive, while complexity-related factors are mostly negative. This suggests that the signals reflect different aspects of model behavior, including cases where the model can overcome apparent difficulty and cases where perceived difficulty corresponds to actual failure.

Together, the two signals are stronger in terms of accuracy because they solve complementary cases. Trust-only routing can miss examples that look reliable but are structurally difficult, while complexity-only routing can over-switch examples that look difficult but are still answered correctly in text. The full adaptive policy combines evidence reliability with instance difficulty, giving a more stable accuracy-cost tradeoff than either signal alone. Overall, adaptive switching should be interpreted as a helpful token-reduction method in some settings and as a mechanism for allocating expensive structured reasoning to examples where the model is more likely to need it.
\begin{table*}[t]
\centering
\small
\setlength{\tabcolsep}{2.4pt}
\renewcommand{\arraystretch}{0.95}
\begin{tabular}{llccc}
\toprule
\textbf{Dataset} & \textbf{Component} & \textbf{Qwen3-32B} & \textbf{LLaMA3.1-70B} & \textbf{GPT-5.1} \\
\midrule
StepGame & Thresholds 
& $\tau_t{=}0.95,\tau_c{=}0.50$ 
& $\tau_t{=}0.75,\tau_c{=}0.50$ 
& $\tau_t{=}0.75,\tau_c{=}0.40$ \\
\cmidrule(lr){2-5}
StepGame & Text & 2,012 & 272 & 249 \\
StepGame & Grid pipeline & 19,218 & 15,472 & 15,727 \\
\cmidrule(lr){2-5}
StepGame & Trust signal only & 4,545 & 2,871 & 3,541 \\
StepGame & \quad Trust-only acc. & 83.6 & 78.1 & 81.8 \\
StepGame & Complexity signal only & 5,792 & 4,622 & 4,696 \\
StepGame & \quad Complexity-only acc. & 73.8 & 71.7 & 84.0 \\
\cmidrule(lr){2-5}
StepGame & Switch signal & 6,440 & 3,998 & 4,845 \\
StepGame & Switch total & 19,137 & 15,928 & 17,399 \\
StepGame & Switch rate & 69.6 & 78.8 & 80.40 \\
StepGame & Adaptive accuracy & 84.4 & 83.6 & 85.2 \\
\midrule
SpaRTUN & Thresholds 
& $\tau_t{=}0.80,\tau_c{=}0.50$ 
& $\tau_t{=}0.95,\tau_c{=}0.60$
& $\tau_t{=}0.95,\tau_c{=}0.50$ \\
\cmidrule(lr){2-5}
SpaRTUN & Text & 1,973 & 1,122 & 1,145 \\
SpaRTUN & Grid pipeline & 28,113 & 31,171 & 26,951 \\
\cmidrule(lr){2-5}
SpaRTUN & Trust signal only & 10,021 & 5,637 & 4,891 \\
SpaRTUN & \quad Trust-only acc. & 69.9 & 61.2 & 75.3 \\
SpaRTUN & Complexity signal only & 2,961 & 2,968 & 1,173 \\
SpaRTUN & \quad Complexity-only acc. & 70.8 & 56.9 & 77.0 \\
\cmidrule(lr){2-5}
SpaRTUN & Switch signal & 9,671 & 6,419 & 4,543 \\
SpaRTUN & Switch total & 26,185 & 29,191 & 23,724 \\
SpaRTUN & Switch rate & 55.5 & 72.3 & 69.9 \\
SpaRTUN & Adaptive accuracy & 74.50 & 60.80 & 77.40 \\
\bottomrule
\end{tabular}%
\caption{Average token cost and routing performance for adaptive switching. Text reports the average token cost of text-only reasoning. Grid pipeline reports the average token cost of the full grid-based reasoning route. Trust signal only and Complexity signal only report the average cost of computing only the trustworthiness or complexity signal, respectively; their corresponding accuracy rows report the final routing accuracy when the policy uses only that signal. Switch signal reports the average token cost of computing the full adaptive switching decision using both trustworthiness and complexity, with early stopping for the plausibility computation when the faithfulness score is already sufficient to determine the routing decision. Switch total reports the total average token cost after adding the switching signal cost to the selected reasoning route. Switch rate is the percentage of examples routed to the grid pipeline. Adaptive accuracy is the final exact-match accuracy of the full switching policy. Thresholds $(\tau_t,\tau_c)$ are tuned on the validation split and evaluated on the test split.}
\label{tab:routing_costs}
\end{table*}

\section{Coordinate Layout} 
\label{app:prolog-coordinate} 
For the coordinate-based representation, we formulate the extracted spatial relations as a constraint-satisfaction problem using Prolog's Constraint Logic Programming over Finite Domains (CLP(FD))~\citep{prolog,clpfd}. Each entity $E$ is associated with coordinate variables $(x_E,y_E)$ through the helper predicate \texttt{xy/3}, and spatial predicates impose constraints over these variables. For example, the relation \texttt{left(A,B)} requires $A$ and $B$ to lie on the same row and places $A$ one unit to the left of $B$: \begin{verbatim} left(A,B) :- xy(A,X1,Y1), xy(B,X2,Y2),
 Y1 #= Y2, X1 + 1 #= X2. \end{verbatim}
 Inverse relations are defined by reversing the predicate arguments, for example, \begin{verbatim} right(A,B) :- left(B,A). \end{verbatim} 
 Other StepGame directional relations are translated analogously into constraints over the horizontal and vertical coordinates. \paragraph{From relation triples to Prolog constraints.} The relation-extraction stage produces triples of the form \texttt{\{head, relation, tail\}}. Each triple is deterministically mapped to its corresponding Prolog predicate. For example, \begin{verbatim} {head: x, relation: left, tail: w} 
 {head: o, relation: right, tail: w} \end{verbatim}
 is translated into \begin{verbatim} left(x,w), right(o,w). \end{verbatim}
 All predicates extracted from a story are solved jointly rather than independently. For the example above, Python submits a query of the form \begin{verbatim} left(x,w), right(o,w), 
 flatten_vars(Vars),
 labeling([ffc,bisect], Vars),
 xy(x,X1,Y1), xy(w,X2,Y2), xy(o,X3,Y3). \end{verbatim}
 Here, \texttt{flatten\_vars} collects the coordinate variables and \texttt{labeling} searches for values satisfying the complete set of spatial constraints. Prolog may return a satisfying assignment such as \begin{verbatim} X1 = 0, Y1 = 0, X2 = 1, Y2 = 0, X3 = 2, Y3 = 0. \end{verbatim}
 which is mapped by Python to \begin{verbatim} X = (0,0), W = (1,0), O = (2,0). \end{verbatim}
 Thus, $X$ is positioned one unit to the left of $W$, while $O$ is one unit to its right. If the extracted constraints are mutually inconsistent, Prolog returns no satisfying assignment. The resulting coordinate assignment is serialized with the question and provided to the language model for coordinate-based reasoning. \paragraph{Why coordinate layouts are limited to StepGame.} We apply the Prolog-based coordinate representation only to StepGame. Its pairwise directional relations can still admit a coordinate assignment when some extracted relations are missing or incorrect, although the resulting layout may be incomplete or inaccurate. In SpaRTUN and ReSQ, richer and more interdependent relations such as containment, topology, depth, and distance make coordinate layouts substantially more sensitive to extraction errors. We therefore use 2D grid representations with these richer spatial relations encoded symbolically rather than Prolog-based coordinates for these datasets.

\section{Error Analysis}
\label{app:error_taxonomy}

This appendix analyzes the main failure modes behind text-only reasoning, grid-based reasoning, and adaptive switching. We focus on three diagnostic views: the error taxonomy, disagreements between text and grid routes, and how often switching recovers different categories of text-only failures.

\subsection{Error Taxonomy}
\label{app:taxonomy_definition}

We use error analysis as a diagnostic tool to understand when switching from text-only reasoning to grid-based reasoning helps, and when it can still fail. This analysis is separate from the larger switching experiments reported in the main results. It is conducted on a diagnostic subset of the \textsc{StepGame} test set for which we inspect model traces, generated grids, and gold grids.

We classify errors at two levels. \textbf{Input-level labels} describe structural properties of the story that make an instance difficult independent of the model's behavior. \textbf{Output-level labels} describe the specific failure mode exhibited by the model. Since a single example may involve both a difficult input structure and an incorrect model behavior, labels are not mutually exclusive.

\textbf{Input-level labels.}
\emph{Composite spatial} marks stories that contain diagonal or composite directional relations, such as \texttt{upper-right} or clock-face expressions like ``at the 7 o'clock position.'' These cases require decomposing a relation across two spatial axes.
\emph{Multi-hop} marks stories where the question cannot be answered from a single stated relation and instead requires chaining two or more relations across sentences. For example, the model may need to invert ``H is right of Z'' into ``Z is left of H'' and then compose it with another relation such as ``H is upper-right of G.''
\begin{table*}
\centering
\small
\setlength{\tabcolsep}{6pt}
\begin{tabular}{lccccc}
\toprule
\textbf{Model} & \textbf{$n$} & \textbf{Text$\checkmark$ Grid$\times$} & \textbf{Grid$\checkmark$ Text$\times$} & \textbf{Both$\checkmark$} & \textbf{Both$\times$} \\
\midrule
Qwen3-32B    & 250 & 20  & 66  & 143 & 21 \\
Qwen3-8B     & 250 & 9   & 103 & 105 & 33 \\
Qwen3-14B    & 250 & 24  & 96  & 112 & 18 \\
GPT-5.1      & 250 & 20  & 51  & 158 & 21 \\
LLaMA3.1-8B  & 250 & 7   & 125 & 67  & 51 \\
LLaMA3.1-70B & 250 & 3   & 111 & 103 & 33 \\
\bottomrule
\end{tabular}
\caption{
Agreement and disagreement between text-only and grid-based reasoning on the diagnostic \textsc{StepGame} subset.
\textbf{$n$} is the number of evaluated instances.
\textbf{Text$\checkmark$ Grid$\times$} counts cases where text-only reasoning is correct but grid-based reasoning is incorrect.
\textbf{Grid$\checkmark$ Text$\times$} counts cases where grid-based reasoning is correct but text-only reasoning is incorrect.
\textbf{Both$\checkmark$} counts cases where both routes are correct.
\textbf{Both$\times$} counts cases where both routes fail.
The large number of \textbf{Grid$\checkmark$ Text$\times$} cases shows that grids often recover text-only failures, while the non-zero \textbf{Text$\checkmark$ Grid$\times$} column shows that grids can also introduce errors. This motivates adaptive switching rather than always using the grid.
}
\label{tab:text_grid_disagreement_stepgame}
\end{table*}
\textbf{Output-level labels.}
For text-only failures, we assign one or more of four labels using a GPT-based classifier that observes the story, question, gold answer, and sentence-level reasoning trace. \emph{Multi-hop reasoning failure} is assigned when the model has access to the necessary relations but fails to compose them correctly. \emph{Composite failure} is assigned when the gold answer is diagonal but the model collapses it to a single axis, for example predicting \texttt{below} when the gold answer is \texttt{lower-left}. \emph{Linguistic difficulty} is assigned when the model misinterprets non-canonical spatial language, including clock-face references, informal directional phrases, or unusual prepositions. \emph{Hallucination} is assigned when the reasoning trace introduces a relation that is not grounded in the story.

For grid failures, we use a parallel classifier that also observes the generated grid and the gold-standard grid. We attribute each grid failure to either \textbf{relation-extraction failure}, where the grid is incorrectly constructed because the extracted relations do not match the gold relations, or \textbf{grid-reasoning failure}, where the grid is correctly constructed but the model misreads or fails to reason well over the layout.

\subsection{Reasoning Modality Disagreement}
\label{app:modality_disagreement}

Before analyzing individual error types, we first compare whether text-only and grid-based reasoning succeed on the same subset of examples. This helps separate two effects: cases where the grid recovers an error made by text-only reasoning and cases where the grid introduces a new error even though text-only reasoning was correct. This disagreement analysis is computed on the same diagnostic \textsc{StepGame} subset and is intended to explain the behavior of the switching policy rather than replace the larger-scale switching results in the main experiments.

Table~\ref{tab:text_grid_disagreement_stepgame} shows that grid-based reasoning frequently corrects text-only failures across models. This effect is especially visible for smaller or weaker models, where \textbf{Grid$\checkmark$ Text$\times$} is much larger than \textbf{Text$\checkmark$ Grid$\times$}. However, the grid is not universally beneficial. In some cases, text-only reasoning gives the correct answer while the grid route fails, either because relation extraction produces an imperfect grid or because the model misreads the generated layout. These disagreements support the need for a selective switching policy rather than a fixed choice of reasoning modality.

\begin{table*}
\centering
\scriptsize
\setlength{\tabcolsep}{3pt}
\renewcommand{\arraystretch}{0.92}
\begin{tabular}{lllrrrrrrr}
\toprule
\textbf{Model} & \textbf{Level} & \textbf{Err./Diff. Type}
    & \textbf{\%Fail} & \textbf{Sw.} & \textbf{No Sw.}
    & \textbf{Rec.} & \textbf{Grid Fail} & \textbf{RE Fail} & \textbf{Grid Rsn.} \\
\midrule

\multicolumn{10}{l}{\textit{GPT-5.1 ($n{=}250$; 72 text-only failures; 85.2\% final adaptive-switch accuracy)}} \\
\midrule
GPT-5.1 & Input  & Composite spatial     & 100.0 & 91.7 &  8.3 & 71.2 & 28.8 & 84.2 & 15.8 \\
GPT-5.1 & Input  & Multi-hop             &  79.2 & 94.7 &  5.3 & 66.7 & 33.3 & 83.3 & 16.7 \\
GPT-5.1 & Output & Multi-hop failure     &  54.2 & 92.3 &  7.7 & 80.6 & 19.4 & 100.0 & --- \\
GPT-5.1 & Output & Linguistic difficulty &  62.5 & 93.3 &  6.7 & 59.5 & 40.5 & 82.4 & 17.6 \\
GPT-5.1 & Output & Hallucination         &  23.6 & 82.4 & 17.6 & 57.1 & 42.9 & 83.3 & 16.7 \\
GPT-5.1 & Output & Composite failure     &  23.6 & 82.4 & 17.6 & 42.9 & 57.1 & 100.0 & --- \\

\midrule
\multicolumn{10}{l}{\textit{Qwen3-32B ($n{=}250$; 87 text-only failures; 84.4\% final adaptive-switch accuracy)}} \\
\midrule
Qwen3-32B & Input  & Composite spatial     & 97.7 & 94.1 &  5.9 & 75.0 & 25.0 & 65.0 & 35.0 \\
Qwen3-32B & Input  & Multi-hop             & 78.2 & 92.6 &  7.4 & 76.2 & 23.8 & 66.7 & 33.3 \\
Qwen3-32B & Output & Multi-hop failure     & 60.9 & 96.2 &  3.8 & 82.4 & 17.6 & 44.4 & 55.6 \\
Qwen3-32B & Output & Linguistic difficulty & 52.9 & 97.8 &  2.2 & 71.1 & 28.9 & 76.9 & 23.1 \\
Qwen3-32B & Output & Hallucination         & 17.2 & 93.3 &  6.7 & 64.3 & 35.7 & 100.0 & --- \\
Qwen3-32B & Output & Composite failure     & 19.5 & 88.2 & 11.8 & 40.0 & 60.0 & 77.8 & 22.2 \\

\midrule
\multicolumn{10}{l}{\textit{LLaMA3.1-70B ($n{=}250$; 144 text-only failures; 84.0\% final adaptive-switch accuracy)}} \\
\midrule
LLaMA3.1-70B & Input  & Composite spatial     & 95.8 & 81.2 & 18.8 & 75.0 & 25.0 & 85.7 & 14.3 \\
LLaMA3.1-70B & Input  & Multi-hop             & 72.9 & 80.0 & 20.0 & 78.6 & 21.4 & 88.9 & 11.1 \\
LLaMA3.1-70B & Output & Multi-hop failure     & 59.7 & 77.9 & 22.1 & 83.6 & 16.4 & 72.7 & 27.3 \\
LLaMA3.1-70B & Output & Linguistic difficulty & 45.8 & 87.9 & 12.1 & 60.3 & 39.7 & 95.7 &  4.3 \\
LLaMA3.1-70B & Output & Hallucination         & 17.4 & 88.0 & 12.0 & 95.5 &  4.5 & 100.0 & --- \\
LLaMA3.1-70B & Output & Composite failure     & 18.8 & 81.5 & 18.5 & 36.4 & 63.6 & 92.9 &  7.1 \\

\bottomrule
\end{tabular}
\caption{
Diagnostic error-based switching analysis for text-only failures on the \textsc{StepGame} subset for stronger models.
Each model block reports the number of evaluated instances, the number of text-only failures analyzed, and the final adaptive-switch accuracy on this subset.
\textbf{Level} indicates whether the label describes an input-level story difficulty or an output-level model failure.
\textbf{Err./Diff. Type} gives the specific difficulty or failure type.
\textbf{\%Fail} is the percentage of text-only failures assigned that label.
\textbf{Sw.} and \textbf{No Sw.} are the percentages of labeled cases that the policy switches or does not switch to the grid route.
\textbf{Rec.} is the percentage of switched cases that are recovered after switching.
\textbf{Grid Fail} is the percentage of switched cases that remain incorrect after switching.
\textbf{RE Fail} is the percentage of remaining grid failures caused by relation-extraction errors.
\textbf{Grid Rsn.} is the percentage of remaining grid failures caused by grid-reasoning errors after grid construction.
Dashes indicate that no cases fall into that category or that the corresponding failure type was not observed.
Labels are not mutually exclusive, so percentages should not be summed across rows.
}
\label{tab:stepgame_switching_stronger_diagnostic}
\end{table*}

\begin{table*}
\centering
\scriptsize
\setlength{\tabcolsep}{3pt}
\renewcommand{\arraystretch}{0.92}
\begin{tabular}{lllrrrrrrr}
\toprule
\textbf{Model} & \textbf{Level} & \textbf{Err./Diff. Type}
    & \textbf{\%Fail} & \textbf{Sw.} & \textbf{No Sw.}
    & \textbf{Rec.} & \textbf{Grid Fail} & \textbf{RE Fail} & \textbf{Grid Rsn.} \\
\midrule

\multicolumn{10}{l}{\textit{Qwen3-14B ($n{=}250$; 114 text-only failures; 88.0\% final adaptive-switch accuracy)}} \\
\midrule
Qwen3-14B & Input  & Composite spatial     & 98.2 & 95.5 &  4.5 & 82.2 & 17.8 & 78.9 & 21.1 \\
Qwen3-14B & Input  & Multi-hop             & 74.6 & 97.6 &  2.4 & 84.3 & 15.7 & 69.2 & 30.8 \\
Qwen3-14B & Output & Multi-hop failure     & 71.1 & 97.5 &  2.5 & 82.3 & 17.7 & 71.4 & 28.6 \\
Qwen3-14B & Output & Linguistic difficulty & 64.0 & 95.9 &  4.1 & 74.3 & 25.7 & 83.3 & 16.7 \\
Qwen3-14B & Output & Hallucination         & 14.0 & 87.5 & 12.5 & 85.7 & 14.3 & 100.0 & --- \\
Qwen3-14B & Output & Composite failure     & 18.4 & 100.0 & --- & 57.1 & 42.9 & 100.0 & --- \\

\midrule
\multicolumn{10}{l}{\textit{LLaMA3.1-8B ($n{=}250$; 176 text-only failures; 36.8\% adaptive accuracy; 10.2\% failure recovery)}} \\
\midrule
LLaMA3.1-8B & Input  & Composite spatial            & 95.5 & 100.0 & --- & 70.2 & 29.8 & 100.0 & --- \\
LLaMA3.1-8B & Input  & Multi-hop reasoning          & 75.0 &  98.5 & 1.5 & 69.2 & 30.8 & 100.0 & --- \\
LLaMA3.1-8B & Output & Multi-hop reasoning failure  & 44.3 &  97.4 & 2.6 & 82.9 & 17.1 & 100.0 & --- \\
LLaMA3.1-8B & Output & Linguistic difficulty        & 50.6 & 100.0 & --- & 55.1 & 44.9 & 100.0 & --- \\
LLaMA3.1-8B & Output & Hallucination                & 35.8 &  98.4 & 1.6 & 53.2 & 46.8 & 100.0 & --- \\
LLaMA3.1-8B & Output & Composite failure            & 19.9 & 100.0 & --- & 48.6 & 51.4 & 100.0 & --- \\

\midrule
\multicolumn{10}{l}{\textit{Qwen3-8B ($n{=}250$; 136 text-only failures; 85.8\% final adaptive-switch accuracy)}} \\
\midrule
Qwen3-8B & Input  & Composite spatial            & 97.1 & 89.4 & 10.6 & 72.9 & 27.1 & 100.0 & --- \\
Qwen3-8B & Input  & Multi-hop reasoning          & 80.9 & 91.8 &  8.2 & 74.3 & 25.7 & 100.0 & --- \\
Qwen3-8B & Output & Multi-hop reasoning failure  & 52.2 & 91.5 &  8.5 & 83.1 & 16.9 & 100.0 & --- \\
Qwen3-8B & Output & Linguistic difficulty        & 50.0 & 94.1 &  5.9 & 57.8 & 42.2 & 100.0 & --- \\
Qwen3-8B & Output & Hallucination                & 15.4 & 81.0 & 19.0 & 58.8 & 41.2 & 100.0 & --- \\
Qwen3-8B & Output & Composite failure            & 23.5 & 96.9 &  3.1 & 51.6 & 48.4 & 100.0 & --- \\

\bottomrule
\end{tabular}
\caption{
Diagnostic error-based switching analysis for text-only failures on the \textsc{StepGame} subset for smaller/open models.
Each model block reports the number of evaluated instances, the number of text-only failures analyzed, and the final adaptive-switch accuracy on this subset.
\textbf{Level} indicates whether the label describes an input-level story difficulty or an output-level model failure.
\textbf{Err./Diff. Type} gives the specific difficulty or failure type.
\textbf{\%Fail} is the percentage of text-only failures assigned that label.
\textbf{Sw.} and \textbf{No Sw.} are the percentages of labeled cases that the policy switches or does not switch to the grid route.
\textbf{Rec.} is the percentage of switched cases that are recovered after switching.
\textbf{Grid Fail} is the percentage of switched cases that remain incorrect after switching.
\textbf{RE Fail} is the percentage of remaining grid failures caused by relation-extraction errors.
\textbf{Grid Rsn.} is the percentage of remaining grid failures caused by grid-reasoning errors after grid construction.
Dashes indicate that no cases fall into that category or that the corresponding failure type was not observed.
Labels are not mutually exclusive, so percentages should not be summed across rows.
}
\label{tab:stepgame_switching_smaller_diagnostic}
\end{table*}

\subsection{Diagnostic Switching Analysis}
\label{app:switching_analysis}

We next analyze how the switching policy behaves on text-only failures in a diagnostic \textsc{StepGame} subset which is 250 samples selected at random with seed 0 from our evaluation sample of 5000 question-answer pairs for Stepgame. Tables~\ref{tab:stepgame_switching_stronger_diagnostic} and~\ref{tab:stepgame_switching_smaller_diagnostic} report which text-only failures are routed to the grid, which switched cases are recovered, and which failures remain after grid-based reasoning. We split the analysis into stronger models and smaller/open models for readability, while keeping the same error categories and metrics across both tables. For this analysis, we randomly sample 250 instances from the switching test set. We use this subset to keep the evaluation computationally tractable, as computing these diagnostics requires LLM-based analysis with models such as GPT.

Tables~\ref{tab:stepgame_switching_stronger_diagnostic} and~\ref{tab:stepgame_switching_smaller_diagnostic} show that multi-hop reasoning failures are the most consistently recoverable category once the policy routes an instance to the grid. Among stronger models, recovery for multi-hop failures is 80.6\% for GPT-5.1, 82.4\% for Qwen3-32B, and 83.6\% for LLaMA3.1-70B. The same trend appears for the smaller/open models, with 82.3\% recovery for Qwen3-14B, 83.1\% for Qwen3-8B, and 82.9\% for LLaMA3.1-8B among switched multi-hop failure cases. This indicates that grid-based reasoning is particularly helpful when the text-only route fails because it cannot compose spatial relations across multiple steps.

However, the overall benefit of switching still depends on whether the model can construct and use the grid reliably. This is clearest for LLaMA3.1-8B: although switched multi-hop failures are often recovered, the model has many text-only failures overall and reaches only 36.8\% adaptive accuracy on the diagnostic subset, recovering 10.2\% of text-only failures. This suggests that switching frequently is not sufficient when the upstream relation extraction and grid-use pipeline are weak. In contrast, Qwen3-8B and Qwen3-14B achieve much higher adaptive accuracies, showing that smaller models can benefit substantially from switching when their structured pipeline is strong enough.

Composite failures are less consistently recovered than multi-hop failures. Their recovery rates are lower for all models, ranging from 36.4\% for LLaMA3.1-70B to 57.1\% for Qwen3-14B. This suggests that composite or diagonal spatial relations remain difficult because the pipeline must first extract, decompose, and encode the relation correctly before grid reasoning can help.

The residual failure columns further show that relation extraction is the main bottleneck in many switched failures. For Qwen3-8B and LLaMA3.1-8B, all observed remaining grid failures in this diagnostic subset are attributed to relation-extraction errors. For GPT-5.1, Qwen3-32B, Qwen3-14B, and LLaMA3.1-70B, some remaining failures are instead due to grid reasoning, meaning that the grid may be sufficiently informative but the model still misreads the layout or reverses the reference frame. Overall, these diagnostics support the main conclusion that switching is most useful for recoverable multi-hop composition errors, while its ceiling is limited by relation extraction quality and by the model's ability to interpret the constructed grid.

\section{Examples}

\subsection{StepGame}
\label{app:stepgame_examples}

Figures~\ref{fig:grid-success-case} and~\ref{fig:grid-failure-case} show StepGame examples using Qwen3-8B for question answering and relation extraction, with grids constructed deterministically from extracted relations. The examples include a case where grids recover a multi-hop text-only failure and a case where grid reasoning still fails because the model misreads row and column positions.

\begin{figure*}[p]
\centering
\begin{tcolorbox}[
    width=0.95\textwidth,
    colback=gray!12,
    colframe=gray!45,
    arc=6pt,
    boxrule=0.6pt,
    title={Both Grids Recover a Text-Only Multi-Hop Failure},
    fonttitle=\bfseries,
    coltitle=black
]
\small
\textbf{Dataset ID:} \texttt{5\_1103} \hfill \textbf{Hop level:} \(k_{\text{hop}}=5\)

\textbf{Story:}
\emph{P is to the bottom right of V. E is over V. X presents lower left to P. Y is positioned in the top left corner of C. Y is positioned right to X.}

\textbf{Question:} What is the relation of agent E to agent Y?

\textbf{Gold:} \texttt{upper-left}
\hfill
\textbf{Text-only prediction:} \texttt{above} \((\neq\) gold)

\textbf{Relation Extraction:} Correct.

\textbf{Full Grid:}
\begin{verbatim}
       Col(1) Col(2) Col(3)
Row(1)  E      -      -
Row(2)  V      -      -
Row(3)  -      P      -
Row(4)  X      Y      -
Row(5)  -      -      C
\end{verbatim}

\textbf{Pruned Grid:}
\begin{verbatim}
       Col(1) Col(2)
Row(1)  E      -
Row(2)  -      Y
\end{verbatim}

\textbf{Full-Grid Justification:} \texttt{upper-left}  
The full grid places \(E\) at Row(1), Col(1) and \(Y\) at Row(4), Col(2). Since \(E\) is in a smaller row and a smaller column than \(Y\), the model correctly infers that \(E\) is \texttt{upper-left} of \(Y\).

\textbf{Pruned-Grid Justification:} \texttt{upper-left}  
The pruned grid keeps only the queried entities while preserving their relative positions. \(E\) remains above and left of \(Y\), so the same row/column comparison recovers the gold answer.

\textbf{Takeaway:}
This example shows how grid grounding can recover a text-only multi-hop failure. The text route collapses the k= 5-hop instance \(E \rightarrow V \rightarrow P \rightarrow X \rightarrow Y\) into the answer \texttt{above}, missing the horizontal component. Both grid views externalize the composed relation as explicit coordinates, reducing the final inference to a direct row/column comparison. The pruned grid further removes distractor entities without changing the relevant relation.
\end{tcolorbox}

\caption{Success case where text-only reasoning fails on a 5-hop diagonal chain, while both full and pruned grids recover the gold answer \texttt{upper-left} through explicit row/column comparison.}
\label{fig:grid-success-case}
\end{figure*}

\begin{figure*}[p]
\centering
\begin{tcolorbox}[
    width=0.95\textwidth,
    colback=gray!12,
    colframe=gray!45,
    arc=6pt,
    boxrule=0.6pt,
    title={Grid Reasoning Fails Due to Layout Misreading},
    fonttitle=\bfseries,
    coltitle=black
]
\small
\textbf{Dataset ID:} \texttt{6\_6123} \hfill \textbf{Hop level:} \(k_{\text{hop}}=6\)

\textbf{Story:}
\emph{W is over M. U and T are side by side with U to the left and T to the right. U is diagonally below M to the left at a 45 degree angle. X and W are in a horizontal line with X on the left. L is diagonally to the bottom right of R. X is below L.}

\textbf{Question:} What is the relation of agent L to agent T?

\textbf{Gold:} \texttt{upper-left}
\hfill
\textbf{Text-only prediction:} \texttt{upper-right} \((\neq\) gold)

\textbf{Relation Extraction:} Correct.

\textbf{Full Grid:}
\begin{verbatim}
       Col(1) Col(2) Col(3)
Row(1)  R      -      -
Row(2)  -      L      -
Row(3)  -      X      W
Row(4)  -      -      M
Row(5)  -      U      T
\end{verbatim}

\textbf{Pruned Grid:}
\begin{verbatim}
       Col(1) Col(2)
Row(1)  L      -
Row(2)  -      T
\end{verbatim}

\textbf{Full-Grid Justification:} \texttt{above} \((\neq\) gold)  
The full grid correctly places \(L\) at Row(2), Col(2), but the model misreads \(T\)'s column as Col(2) instead of Col(3). This collapses the diagonal relation into a vertical comparison, leading to the incomplete prediction \texttt{above}.

\textbf{Pruned-Grid Justification:} \texttt{lower-left} \((\neq\) gold)  
The pruned grid removes distractors and keeps only \(L\) and \(T\), but the model reverses their row order during interpretation. It therefore treats \(L\) as below \(T\), producing \texttt{lower-left} instead of the correct \texttt{upper-left}.

\textbf{Takeaway:}
This example illustrates that grid construction alone is not sufficient when the model misreads the layout. The full grid fails because the reader misaligns \(T\)'s column in a denser layout, while the pruned grid fails for a different reason: it inverts the vertical order of the two retained entities. Thus, even with correct relation extraction, the grid route can fail when the model's first step is an incorrect interpretation of row or column positions.
\end{tcolorbox}

\caption{Failure case where both grid views are incorrect despite correct relation extraction. The full grid misreads \(T\)'s column, while the pruned grid inverts the row order between the queried entities.}
\label{fig:grid-failure-case}
\end{figure*}

\subsection{SpaRTUN}
\label{app:spartun_examples}

Figures~\ref{fig:spa}, \ref{fig:spa1}, and~\ref{fig:spa2} show SpaRTUN examples using Qwen3-32B for question answering and relation extraction. These examples focus on topological and containment reasoning, where grids encode nested boxes with bracketed structures and RCC8-style relations such as \texttt{tpp}, \texttt{ntpp}, and \texttt{dc}.

\begin{figure*}[p]
\centering
\begin{tcolorbox}[
    colback=gray!12,
    colframe=gray!45,
    arc=6pt,
    boxrule=0.6pt,
    title={Correct Topological Reasoning with Nested Blocks},
    fonttitle=\bfseries,
    coltitle=black
]
\small
\textbf{Dataset ID:} \texttt{3999-0}

\textbf{Story:}  
\emph{There exist two blocks, called HHH and LLL. Block HHH covers a medium grey hexagon. This block covers block LLL. A medium grey hexagon is inside and touching block LLL.}

\textbf{Question:} What is the position of LLL relative to HHH?

\textbf{Gold:} \texttt{['TPP']}

\textbf{Relation Extraction:} Correct.  
All relevant containment relations are extracted, including that \texttt{block LLL} is contained in \texttt{block HHH} and touches its boundary.

\textbf{Pruned Grid:}
\begin{verbatim}
               Col(1)
Row(1)  [block HHH:
Row(2)    medium grey hexagon #1_in(block HHH)   #touch-edge
Row(3)    [block LLL:                            #touch-edge
Row(4)      medium grey hexagon #2_in(block LLL) #touch-edge
Row(5)    ]
Row(6)  ]
\end{verbatim}

\textbf{Full Grid:}
\begin{verbatim}
               Col(1)
Row(1)  [block HHH:
Row(2)    medium grey hexagon #1_in(block HHH)   #touch-edge
Row(3)    [block LLL:                            #touch-edge
Row(4)      medium grey hexagon #2_in(block LLL) #touch-edge
Row(5)    ]
Row(6)  ]
\end{verbatim}

\textbf{Pruned Justification:} \texttt{['tpp']}  
The pruned grid preserves the nested structure by showing \texttt{block LLL} inside \texttt{block HHH}. Since \texttt{block LLL} is contained in \texttt{block HHH} and carries the boundary tag \texttt{\#touch-edge}, the model correctly infers tangential proper-part containment:
\[
\text{block LLL} \in \text{block HHH}
\quad\text{and touches its boundary}
\Rightarrow \texttt{tpp}.
\]

\textbf{Full Justification:} \texttt{['tpp']}  
The full grid encodes the same containment structure. It also represents \texttt{block LLL} as nested inside \texttt{block HHH}, with the \texttt{\#touch-edge} tag indicating boundary contact. Therefore, the full grid also recovers the correct topological relation \texttt{tpp} rather than relying on row-based directional comparison.

\textbf{Takeaway:}  
This example shows that when relation extraction and grid construction preserve nested containment, both full and pruned grids can support correct RCC8-style topological reasoning. The pruned grid does not lose the nesting relation; instead, it retains the relevant containment structure while removing unnecessary context.
\end{tcolorbox}
\caption{A successful topological reasoning case where both full and pruned grids preserve nested containment. Since \texttt{block LLL} is represented inside \texttt{block HHH} with boundary contact, both grid readers correctly output \texttt{TPP}.}
\label{fig:spa}
\end{figure*}

\begin{figure*}[p]
\centering
\begin{tcolorbox}[
    colback=gray!12,
    colframe=gray!45,
    arc=6pt,
    boxrule=0.6pt,
    title={Containment Partially Recovered, but Extra Direction Added},
    fonttitle=\bfseries,
    coltitle=black
]
\small
\textbf{Dataset ID:} 4042-3

\textbf{Story:}  
\emph{Two boxes, called one and two, exist in an image. Box one has box two. A medium green thing is inside and touching box one. Box two covers an orange apple.}

\textbf{Question:} Where is the orange thing relative to box one?

\textbf{Gold:} \texttt{['NTPP']}

\textbf{Relation Extraction:} Correct.  
All 3 relations matched.

\textbf{Pruned Grid:}
\begin{verbatim}
               Col(1)
Row(1)  [box one:
Row(2)    medium green thing_in(box one)   #touch-edge
Row(3)    [box two_in(box one):            #inside-clear
Row(4)      orange apple_in(box two)       #touch-edge
Row(5)    ]
\end{verbatim}

\textbf{Full Grid:}
\begin{verbatim}
               Col(1)
Row(1)  [box one:
Row(2)    medium green thing_in(box one)   #touch-edge
Row(3)    [box two_in(box one):            #inside-clear
Row(4)      orange apple_in(box two)       #touch-edge
Row(5)    ]
Row(6)  ]
\end{verbatim}

% \textbf{Pruned Interpretation:}  
% Box one: medium green thing at Row(2).  
% Box two: orange apple at Row(4), box-level \texttt{\#inside-clear}.  
% The pruned interpretation treats box two as a separate entity and does not explicitly preserve that it is inside box one.

% \textbf{Full Interpretation:}  
% The full interpretation keeps the notation \texttt{box two\_in(box one)}, explicitly preserving the nesting relation.  
% The orange apple is listed at Row(4) with \texttt{\#touch-edge}.

\textbf{Pruned Justification:} \texttt{['below', 'right']}  
Although the pruned grid retains the \texttt{box two\_in(box one)} header,
the model fails to use the nested containment structure compositionally and
instead falls back to comparing box headers and row/column positions. It
therefore predicts the spurious directional relations \texttt{below} and
\texttt{right}, missing the correct topological containment relation.

\textbf{Full Justification:} \texttt{['below', 'ntpp']}  
The full run preserves the nesting relation explicitly through \texttt{box two\_in(box one)} and correctly recovers multi-hop containment:
\[
\text{orange apple} \in \text{box two} \in \text{box one}
\Rightarrow \texttt{ntpp}.
\]
However, because the orange apple is also listed at a lower row with \texttt{\#touch-edge}, the model adds an unsupported directional relation \texttt{below} based on row comparison. Since the gold answer contains only \texttt{['NTPP']}, the full output is still incorrect.

\textbf{Takeaway:}  
The full run recovers the correct topological relation but still fails because it adds an unsupported directional label. The pruned run misses the containment structure entirely and falls back to row-based comparison.
\end{tcolorbox}
\caption{An example of containment partially recovered with extra direction added.}
\label{fig:spa1}
\end{figure*}

\begin{figure*}[p]
\centering
\begin{tcolorbox}[
    colback=gray!12,
    colframe=gray!45,
    arc=6pt,
    boxrule=0.6pt,
    title={Full Grid Misses Containment Due to Extra Objects},
    fonttitle=\bfseries,
    coltitle=black
]
\small
\textbf{Dataset ID:} 293-0

\textbf{Story:}  
\emph{There are three boxes, named one, two, and three. Box one contains a medium yellow apple and covers a small orange melon. [...] Box two covers a big orange melon which is to the south of a medium green watermelon. Box two covers the medium green watermelon. It contains a small orange apple [...]. Box three covers box two. A big green melon and a medium green apple are covered by this box. [...]}

\textbf{Question:} What is the position of the small orange apple relative to box three?

\textbf{Gold:} \texttt{['NTPP']}

\textbf{Relation Extraction:} Incorrect.  
Genuine missing RE: \texttt{(medium yellow apple number two, ntpp, box one)} --- a non-query relation.

\textbf{Pruned Grid:}
\begin{verbatim}
                 Col(1)
Row(1)  [box three:
Row(2)    medium green apple number two   #touch-edge
Row(3)    big green melon                 #touch-edge
Row(4)    [box two_in(box three):         #touch-edge
Row(5)      small orange apple            #inside-clear
Row(6)      medium green watermelon #1    #touch-edge
Row(7)    ]
\end{verbatim}

\textbf{Full Grid:}
\begin{verbatim}
                 Col(1)
Row(1)  [box three:
Row(2)    medium green apple number two   #touch-edge
Row(3)    big green melon                 #touch-edge
Row(4)    [box two_in(box three):         #touch-edge
Row(5)      small orange apple            #inside-clear
Row(6)      medium green watermelon #1    #touch-edge
Row(7)      big orange melon              #touch-edge
Row(8)      medium green watermelon #2    #inside-clear
...         [box one_in(box three/two): ...]
\end{verbatim}

\textbf{Pruned Justification:} \texttt{['ntpp']}  
The pruned grid preserves the relevant nesting chain: the small orange apple is inside \texttt{box two}, and \texttt{box two} is inside \texttt{box three}. Therefore, the model can apply multi-step containment reasoning:
\[
\text{small orange apple} \in \text{box two} \in \text{box three}
\Rightarrow \texttt{ntpp}.
\]
Because the pruned layout is sparse and does not provide a reliable directional comparison, it correctly avoids adding an unsupported directional label.

\textbf{Full Justification:} \texttt{['below']}  
The full grid contains additional boxes and a longer layout, so the model focuses on row positions instead of propagating containment through \texttt{box two} and \texttt{box three}. It treats \texttt{box two} as lower than \texttt{box three} and predicts the directional relation \texttt{below}. This misses the correct topological relation, since the gold answer depends on nested containment rather than row-based comparison.

\textbf{Takeaway:}  
This is a clear case where the pruned run succeeds because it preserves the essential nesting structure and applies multi-step containment:
\[
\text{apple} \in \text{box two} \in \text{box three} \Rightarrow \texttt{ntpp}.
\]
The full run, however, includes additional objects and boxes, making row-order reasoning more salient. As a result, it outputs only \texttt{below} and misses the correct topological relation.
\end{tcolorbox}
\caption{An example where the full grid misses containment due to extra objects.}
\label{fig:spa2}
\end{figure*}

\subsection{ReSQ}
\label{app:resq_examples}

Figures~\ref{fig:resq1} and~\ref{fig:resq2} show qualitative examples from ReSQ using Qwen3-32B for question answering and relation extraction, and GPT-5.1 for grid construction. The grid is represented as a \(5\times5\) markdown table with directional tags such as \texttt{\#above(X)}, \texttt{\#below(X)}, \texttt{\#left(X)}, \texttt{\#right(X)}, \texttt{\#front(X)}, \texttt{\#behind(X)}, and \texttt{\#near(X)}. 

\begin{figure*}[p]
\centering
\begin{tcolorbox}[
    colback=gray!12,
    colframe=gray!45,
    arc=6pt,
    boxrule=0.6pt,
    title={Above/Below Relation Correct in Text and Both Grids},
    fonttitle=\bfseries,
    coltitle=black
]
\small
\textbf{Image ID:} \texttt{1088} \hfill \textbf{Ground Truth:} Yes \hfill
\textbf{Text-based:} Yes \hfill \textbf{Full Grid:} Yes \hfill \textbf{ Grid:} Yes

\textbf{Story:}
\emph{A photo of a room with white walls, two single beds with a night table in
between and a picture on the wall above the beds.}

\textbf{Question:} Are the beds below the picture?

\textbf{Selected Entities (after pruning):}
\texttt{['picture on the wall', 'two single beds']}

\textbf{Full Grid:}
\begin{verbatim}
|          | **Col1**                                  | **Col2**    | **Col3** |
| **Row1** | picture on the wall #above(two single beds), #near(two single beds)
, #near(night table), #front(white walls) | — | — |
| **Row2** | white walls #behind(two single beds), #behind(night table),
#behind(picture on the wall) | — | — |
| **Row3** | two single beds #below(picture on the wall), 
#near(night table), #near(picture on the wall) | night table | — |
\end{verbatim}

\textbf{Pruned Grid:}
\begin{verbatim}
|          | **Col1** | **Col2** | **Col3** |               
| **Row1** |  picture on the wall #above(two single beds) | —      | —        |
| **Row2** | —        | —        | —      |                   
| **Row3** | two single beds #below(picture on the wall), #near(picture on the wall) 
| — | — |
\end{verbatim}

\textbf{Text-Only Justification:} \texttt{Yes} \\
The story states that there is a picture on the wall ``above the beds,'' which directly implies that the beds are below the picture.

\textbf{Full-Grid Justification:} \texttt{Yes} \\
The two single beds are in Row3, Col1 and the picture on the wall is in Row1, Col1. Since row indices increase downward, the beds are below the picture. The tag \texttt{\#below(picture on the wall)} also makes this relation explicit.

\textbf{Pruned-Grid Justification:} \texttt{Yes} \\
The pruned grid preserves the two relevant entities and their vertical relation. The beds remain in Row3 and the picture remains in Row1, with the \texttt{\#below(picture on the wall)} tag attached to the beds.

\textbf{Takeaway:}
This is a clean success case across all three modes. The above/below relation is stated directly in the story, encoded positionally in the full grid, and preserved by pruning. The pruned grid removes distractor entities such as \emph{white walls} and \emph{night table} while retaining the relation needed to answer the question.
\end{tcolorbox}
\caption{An example from ReSQ dataset correctly answered by text-only reasoning and by both grid views.}
\label{fig:resq1}
\end{figure*}

\begin{figure*}[p]
\centering
\begin{tcolorbox}[
    colback=gray!12,
    colframe=gray!45,
    arc=6pt,
    boxrule=0.6pt,
    title={Commonsense Recovery in the Pruned Grid},
    fonttitle=\bfseries,
    coltitle=black
]
\small
\textbf{ID:} \texttt{1145} \hfill \textbf{Ground Truth:} Yes \hfill
\textbf{Text-based :} No \hfill \textbf{Full Grid:} No \hfill \textbf{Pruned Grid:} Yes

\textbf{Story:}
\emph{A courtyard with stairs on the left, a big palm tree in the centre, a few tables and chairs and a light brown house with red rooftiles in the background.}

\textbf{Question:} Is the house below the rooftiles?

\textbf{Selected Entities (after pruning):}
\texttt{['light brown house with red rooftiles']}

\textbf{Full Grid:}
\begin{verbatim}
|          | **Col1**     | **Col2** | **Col3**        | 
| **Row1** | —       |  light brown house with red rooftiles
#behind(stairs), #behind(big palm tree), #behind(tables), #behind(chairs) | — |
| **Row2** | —       |big palm tree #right(stairs), #near(tables), #near(chairs),
#front(light brown house with red rooftiles) | — | 
| **Row3** | stairs #left(big palm tree), #left(tables), 
#left(chairs), #front(light brown house with red rooftiles) | — | — |
| **Row4** | —        | tables #right(stairs), #near(big palm tree),
#near(chairs), #front(light brown house with red rooftiles) | — | 
| **Row5** | —         | chairs #right(stairs), #near(big palm tree),
#near(tables), #front(light brown house with red rooftiles) | — | 
\end{verbatim}

\textbf{Pruned Grid:}
\begin{verbatim}
|          | **Col1** | **Col2** | **Col3**                       
| **Row1** | light brown house with red rooftiles  | —   | —  |
| **Row2** | —        | —        | —       |
| **Row3** | —        | —        | —       |                           

\end{verbatim}

\textbf{Text-Only Justification:} \texttt{No} \\
The text-only route treats ``light brown house with red rooftiles'' as a single compound entity and therefore does not infer a separate above/below relation between the house and the rooftiles.

\textbf{Full-Grid Justification:} \texttt{No} \\
The full grid also represents ``light brown house with red rooftiles'' as a single object in Row1, Col2. Because the house and rooftiles are not separated into distinct grid entities, the model does not derive a vertical relation between them.

\textbf{Pruned-Grid Justification:} \texttt{Yes} \\
The pruned grid contains only the compound entity. With distractor objects removed and no competing row-based comparison, the model relies on commonsense knowledge that rooftiles are located on top of a house, correctly answering that the house is below the rooftiles.

\textbf{Takeaway:}
This example highlights a limitation and a benefit of pruning. The relation is not represented explicitly because the house and rooftiles are fused into one entity mention. Text-only and full-grid reasoning both reject the relation due to this surface-form constraint. The pruned grid removes irrelevant spatial anchors and allows the model to recover the gold answer through commonsense about the structure.
\end{tcolorbox}
\caption{An example from ReSQ dataset incorrectly answered by text-only and full grid reasoning and correctly by pruned grid.}
\label{fig:resq2}
\end{figure*}

\begin{figure*}[p]
\centering
\begin{tcolorbox}[
    width=0.95\textwidth,
    colback=gray!12,
    colframe=gray!45,
    arc=6pt,
    boxrule=0.6pt,
    title={Worked Example of Trust, Complexity, and Switching},
    fonttitle=\bfseries,
    coltitle=black
]
\small

\textbf{Model / thresholds:} Qwen3-8B, $\tau_t=0.85$, $\tau_c=0.45$.
\quad \textbf{StepGame split:} $k_{\text{hop}}=8$.

\textbf{Story:}
\emph{U is placed in the right direction of V. M is positioned in the lower
left corner of A. K is on the right side and top of M. T is to the left of C
with a small gap between them. U is sitting at the 9:00 position of T. E is
positioned in the lower right corner of X. X is below and to the left of C.
The object K is positioned below and to the left of the object E.}

\textbf{Question:} What is the relation of the agent T to the agent E?
\quad
\textbf{Gold:} \texttt{upper-left}

\vspace{0.3em}
\textbf{Faithfulness ($F$).}
The model identifies the relevant support chain through $C$ and $X$:
\begin{verbatim}
"T is to the left of C with a small gap between them."
"X is below and to the left of C."
"E is positioned in the lower right corner of X."
\end{verbatim}
The support-only answer matches the original answer, and removing this support
makes the answer unavailable. Thus, both sufficiency and necessity hold:
\[
S=1,\qquad N=1,\qquad F=\tfrac{1}{2}(S+N)=1.0.
\]
Since $F=1.0$, trust cannot be ruled out from faithfulness alone: the maximum
possible trust is $0.6F+0.4=1.0$.

\vspace{0.3em}
\textbf{Complexity ($C$).}
The relevant support path is
$T \rightarrow C \rightarrow X \rightarrow E$, and the instance contains
diagonal spatial relations. The StepGame complexity components are
\[
SB=0.375,\quad SD=0.625,\quad CL=0.615,\quad
HL=0.300,\quad DB=0.667.
\]
For example, the instance has three supporting sentences and StepGame uses
$c=5$, giving
\[
SB=\frac{3}{3+5}=0.375.
\]
Using the StepGame weights from Table~\ref{tab:c3_weights},
\[
\begin{aligned}
C
&=0.22(SB)+0.17(SD)+0.22(CL)+0.28(HL)+0.11(DB)\\
&=0.22(0.375)+0.17(0.625)+0.22(0.615)
 +0.28(0.300)+0.11(0.667)\\
&=0.481.
\end{aligned}
\]
Since $C=0.481 \geq \tau_c=0.45$, the short-circuit policy switches to
grid-based reasoning at this stage.

\vspace{0.3em}
\textbf{Plausibility ($P$), reported for illustration.}
The paraphrase and flip checks reveal instability. Two paraphrased variants
preserve \texttt{upper-left}, while one simplified variant drops the vertical
component and answers only \texttt{left}. The flipped question should invert
the relation from $E$ to $T$ as \texttt{lower-right}, but the model again
predicts \texttt{upper-left}. Therefore,
\[
PS=\tfrac{2}{3}, \qquad FC=0, \qquad
P=\tfrac{1}{2}(PS+FC)=\tfrac{1}{2}(0.667+0)=0.333.
\]
The corresponding complete trust score is
\[
T=0.6F+0.4P=0.6(1.0)+0.4(0.333)=0.733 < \tau_t.
\]
For illustration, we report the remaining trustworthiness signals even though
the deployed short-circuit policy has already switched after observing
$C\geq\tau_c$.

\vspace{0.3em}
\textbf{Switch Decision.}
\[
\boxed{\text{SWITCH to grid}}
\]

\textbf{Text-only answer:} \texttt{left} \quad ($\neq$ gold)
\qquad
\textbf{Grid answer:} \texttt{upper-left} \quad (= gold)

\vspace{0.3em}
\textbf{Efficiency note.}
Here, $F=1.0$ is not sufficient to determine the route, so complexity is
evaluated next. Because $C\geq\tau_c$, the deployed policy can switch without
computing plausibility; the complete plausibility and trust scores above are
shown only to illustrate that trustworthiness also supports the switch.

\vspace{0.3em}
\textbf{Takeaway.}
This example shows why faithfulness alone is insufficient. The answer appears
grounded in the correct support chain, but the complete signals reveal a
brittle spatial interpretation: one paraphrase loses the diagonal component
and the flipped question fails completely. The instance is also diagonal-heavy
and multi-hop, so both trustworthiness and complexity indicate that text-only
reasoning is unreliable. The grid representation makes the composed relation
explicit and recovers the gold answer \texttt{upper-left}.

\end{tcolorbox}

\caption{Worked StepGame switching example ($k_{\text{hop}}=8$). High
complexity triggers the short-circuit switch to the grid; complete
trustworthiness signals are also shown for illustration. Grid reasoning
recovers the gold answer \texttt{upper-left}.}
\label{fig:switch}
\end{figure*}

\end{document}